%% file: main.tex
\documentclass[runningheads]{llncs}
\usepackage[T1]{fontenc}
\usepackage{graphicx}
\usepackage{hyperref}
\usepackage{color}

\usepackage{amsmath}
\usepackage{amssymb}
\usepackage{mathtools}
\usepackage{enumitem}
\usepackage{multirow}
\usepackage{multicol}
\usepackage{booktabs}
\newcommand{\modelname}{HypKG}
\usepackage{algorithm}
\usepackage{algorithmic}
\usepackage[toc,title]{appendix}

\begin{document}
\title{HypKG: Hypergraph-based Knowledge Graph Contextualization for Precision Healthcare}
\titlerunning{HypKG: Hypergraph KG Contextualization for Healthcare}

\author{Yuzhang Xie \and
Xu Han\and
Ran Xu\and
Xiao Hu\and
Jiaying Lu\and
Carl Yang
}

\institute{Emory University, Atlanta, GA, USA \\
\email{\{yuzhang.xie,andy.han,ran.xu,xiao.hu,jiaying.lu,j.carlyang\}@emory.edu}\\}
\authorrunning{Y. Xie et al.}
%
%
\maketitle              
\input{sections/1_abstract}
\input{sections/2_introduction}
\input{sections/3_relatedwork}
\input{sections/4_method}

\input{sections/5_experiment}
\input{sections/6_conclusion}
\input{sections/7_acknowledgement}
\newpage
\bibliographystyle{splncs04}
\bibliography{main}
\include{supplementary}
\end{document}

%% file: sections/1_abstract.tex
\begin{abstract}
Knowledge graphs (KGs) are important products of the semantic web, which are widely used in various application domains. Healthcare is one of such domains where KGs are intensively used, due to the high requirement for knowledge accuracy and interconnected nature of healthcare data.
However, KGs storing general factual information often lack the ability to account for important contexts of the knowledge such as the status of specific patients, which are crucial in precision healthcare.
Meanwhile, electronic health records (EHRs) provide rich personal data, including various diagnoses and medications, which provide natural contexts for general KGs.
In this paper, we propose \modelname, a framework that integrates patient information from EHRs into KGs to generate contextualized knowledge representations for accurate healthcare predictions. Using advanced entity-linking techniques, we connect relevant knowledge from general KGs with patient information from EHRs, and then utilize a hypergraph model to ``contextualize'' the knowledge with the patient information. Finally, we employ hypergraph transformers guided by downstream prediction tasks to jointly learn proper contextualized representations for both KGs and patients, fully leveraging existing knowledge in KGs and patient contexts in EHRs.
In experiments using a large biomedical KG and two real-world EHR datasets, \modelname\ demonstrates significant improvements in healthcare prediction tasks across multiple evaluation metrics. 
Additionally, by integrating external contexts, \modelname\ can learn to adjust the representations of entities and relations in KG, potentially improving the quality and real-world utility of knowledge. 
\keywords{Contextualized Knowledge graph  \and Hypergraph \and Knowledge graph representation \and Healthcare application of semantic webs.}
\end{abstract}

%% file: sections/2_introduction.tex
\section{Introduction}
Knowledge graphs (KGs) are important products of the semantic web, organizing and linking entities such as people, places, and objects along with their relationships, in graph structures \cite{kejriwal2022knowledge}. They provide a powerful method for representing factual knowledge about the real world \cite{yang2022rethinking}\cite{cao2024knowledge}, and have found broad applications across various domains. Healthcare is one of the domains where KGs are extensively utilized, driven by the need for highly accurate knowledge and the inherently interconnected nature of healthcare data \cite{abu2023hkg}. In the healthcare domain, KGs capture specialized knowledge about medical concepts, including drugs, diseases, genes, and symptoms \cite{zhang2020hkg}. In recent years, there has been significant growth in the development of large-scale healthcare KGs, such as UMLS \cite{bodenreider2004umls}, Concept5.5 \cite{speer2017conceptnet}, and iBKH \cite{su2023ibkh}. These healthcare KGs provide rich factual knowledge, enhancing clinical research, decision-making, and healthcare delivery \cite{cui2023hkg}.
 
However, KGs storing general facts often fail to account for important contextual information about the knowledge, such as patients' statuses  (e.g., demographics, medical conditions, lifestyles), which can lead to inaccuracies in the knowledge they provide. To ensure the accuracy and relevance of knowledge, it is essential to consider contexts from other data sources. In the healthcare domain, a natural consideration is to utilize KGs together with patient-specific contexts, which is crucial for precision healthcare. 
For instance, a patient's health status can significantly affect the relevance of drug-disease relationships retrieved from a KG. While a general KG may indicate that aspirin can be used for heart disease prevention \cite{gasparyan2008aspirin}, this recommendation may be inappropriate for patients with conditions such as gastrointestinal issues or bleeding disorders, where aspirin could cause adverse effects \cite{de2012aspirin}. Another example involves the patient's medication context. A KG might provide information about commonly prescribed medications for depression, but these recommendations may not be suitable for patients taking other medications that could interact with antidepressants \cite{wolff2021antidepressants}. Without incorporating data about a patient's existing prescriptions, the KG might suggest treatments that could result in harmful drug interactions. 

In the healthcare domain, a natural way to provide context for knowledge in KGs is to use patient information such as data from Electronic Health Records (EHRs). EHRs are widely recognized as valuable assets for comprehensive patient health analysis and informed decision-making \cite{xu2022CACHE}. They are usually constructed from a wide range of patients' digital medical information, including tabular data, clinical notes, medical images, genomics data, and other essential patient data types \cite{abul2019EHR}\cite{xie2024improving}\cite{sun2018EHR}\cite{xie2022survival}\cite{wu2025prediction}. By consolidating this diverse data, EHRs facilitate a deep understanding of patients' health statuses, enabling more precise medical interventions. Moreover, EHR data often include rich contextual details, such as diagnosis, prescriptions, medication histories, admission and discharge times, and even demographic information like age and gender \cite{murali2023towards}.

\begin{figure}[htbp]
\centering
\includegraphics[width=0.8\linewidth]{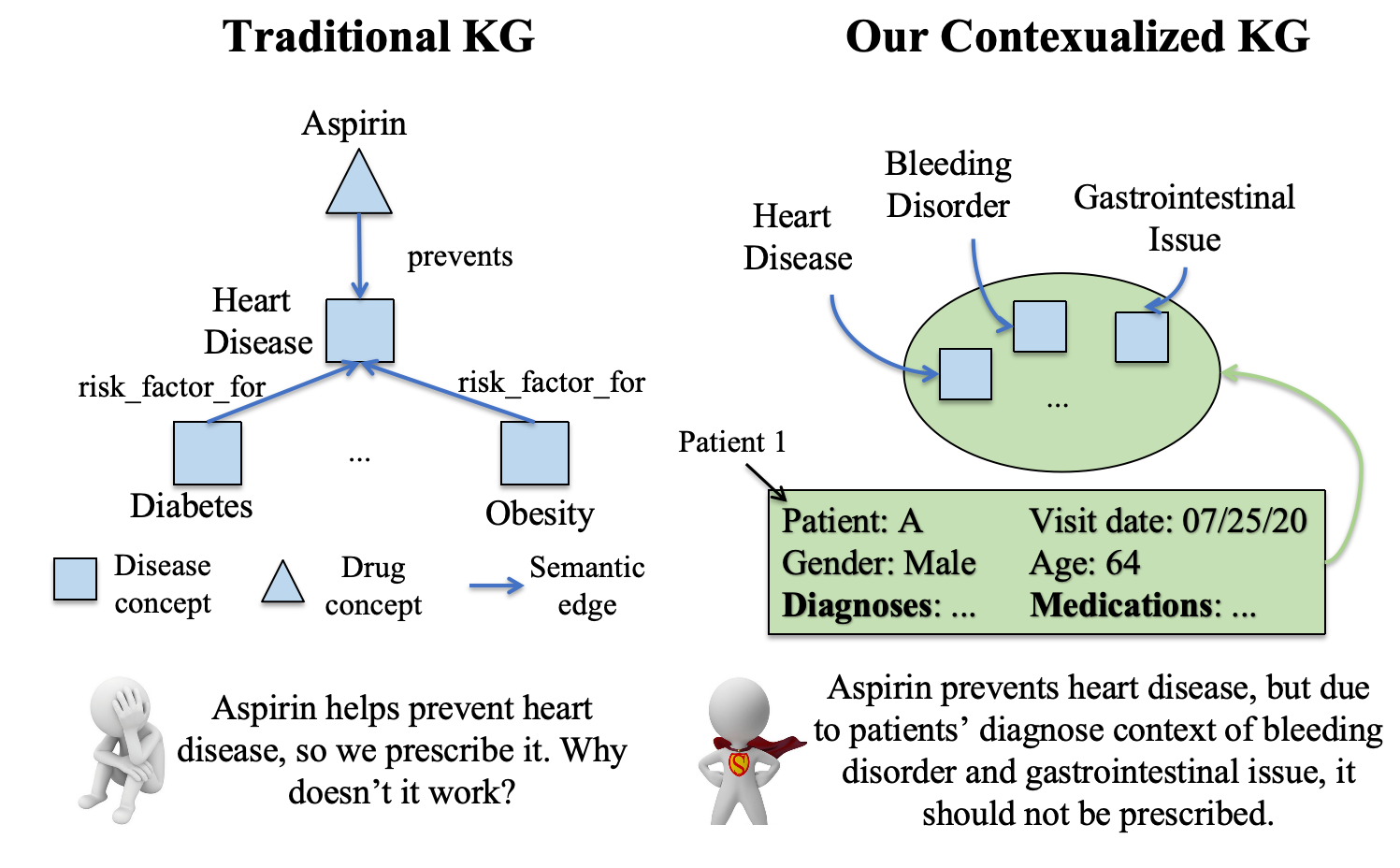}\vspace{-3ex}
\caption{A toy example of KG contextualization. \textit{Left:} traditional KG. \textit{Right:} our proposed contextualized KG.}
\label{fig:toy_example}
\end{figure}

Motivated by the lack of context modeling in KGs for real-world applications like healthcare, and the rich contextual information found in EHRs,
we propose {\modelname}, a novel framework to integrate patients' context information from EHR with extracted knowledge from KG, as exemplified in Figure \ref{fig:toy_example}. 
On one hand, the background knowledge extracted from KGs can serve as a valuable complement to the detailed patient data found in EHRs. This background knowledge can provide general medical information, such as common drug-disease interactions and treatment guidelines, which EHR alone may not explicitly capture. On the other hand, the rich context provided by EHRs, such as patient diagnoses, medication histories, lab results, and personal attributes, can inform the selection of the most relevant knowledge from large-scale KGs, tailoring it to the individual patient's specific needs. This integration of EHR and KG enriches the existing knowledge with patient-specific contexts, which could serve as a foundation for a wide range of downstream healthcare applications.

To combine KG knowledge with relevant context information, several challenges must be addressed. The first challenge is \textbf{how to connect knowledge and external contexts}. For example, there may be discrepancies in the entity naming conventions used in different sources  \cite{kohane2021EL}-- a KG may refer to a disease as ``Ellis-Van Creveld syndrome'', while an EHR may refer to the same disease as ``Chondroectodermal dysplasia''. To address this challenge of biomedical entity linking, researchers have proposed many methods using crafted string-matching rules \cite{d2015ELRule}, constructed thesauri \cite{aronson2010ELDICT}, and pre-trained language models \cite{wang2023ELLM}. In our approach, we leverage the prior knowledge and semantic understanding embedded in advanced \textit{language models} to facilitate the linking process. Through this process, we robustly enable the KG to be linked with patient-specific context information from the EHR.

Another noteworthy challenge lies in \textbf{how to jointly model knowledge and contexts}.
While researchers have proposed various models for EHR data, including feature engineering \cite{yu2021EHRFeature}, deep neural networks (DNNs) \cite{choi2016EHRFeature}, and graph neural networks (GNNs) \cite{ochoa2022EHRFeature}\cite{liu2025graph}, these methods cannot be directly applied to incorporate additional relational knowledge from KGs. 
In this work, we employ a more flexible and unified graph-based data structure of \textit{hypergraphs} \cite{xu2023hypergraph}\cite{zhang2024tacco}, where we model EHR attributes and KG entities with nodes and model patient contexts with hyperedges. Unlike standard graphs where edges connect two nodes, hypergraphs allow hyperedges to connect multiple nodes simultaneously, enabling us to directly model the patient-attribute (context-entity) relationships \cite{johnson2024hypergraph}\cite{lee2024hypergraph}. Additionally, the many-to-many connections in hypergraphs also allow flexible capturing of entity-entity relationships within each patient context, as well as the context-context relationships for each entity. 

The third challenge is \textbf{how to learn a model that properly integrates knowledge and contexts}. 
The foundation of this process involves jointly generating entity representations (embeddings) from large-scale KGs and context (patient) representations from EHRs. For KGs alone, many KG embedding methods have been proposed \cite{yang2022KG}, popular ones including TransE \cite{bordes2011TransE}, ComplEx \cite{trouillon2016ComplEx}, and CompGCN \cite{vashishth2020CompGCN}. These techniques compute compact embedding vectors that capture the entity relationships. We utilize the extensively studied KG embedding methods to capture knowledge in the original KG, and use them to initialize the node features in the hypergraph model.  
Then we incorporate patient context into the model by utilizing \textit{hypergraph transformers} \cite{chien2021settransformer} to directly model the explicit entity-context relationships, and fully learn the entity-entity and context-context relationships from EHR data. We utilize \textit{downstream prediction tasks }to guide the whole training process of hypergraph transformers.
This approach enables the KG information to be combined with the rich, patient-specific context from EHRs. The final outcomes are patient-specific representations as well as contextualized KG representations that integrate the strengths of both data sources.

Overall, the core novelty of \modelname\ lies in its unified, hypergraph‑based framework that integrates patient‑specific contexts from EHRs directly into a general biomedical KG, rather than treating KG embeddings and EHR features separately or constructing isolated ``personalized'' KGs. Unlike prior related work that either (a) embeds KGs without patient-specific contextualization, (b) uses only one‑hot or simple pooled KG vectors for patients, or (c) builds small per‑user graphs, \modelname\ (1) leverages an LLM‑augmented entity‑linking pipeline to align EHR medical attributes with KG entities; (2) represents both KG concepts and patient visits as nodes and hyperedges in a single hypergraph, capturing high‑order relations; and (3) employs hypergraph transformers supervised by downstream healthcare tasks to jointly learn contextualized embeddings. 

In our experiments, we utilize the large-scale public KG, iBKH \cite{su2023ibkh}, as the foundational KG dataset. We enhance this KG by contextualizing it with patient-specific context information from two widely used EHR datasets: MIMIC-III \cite{johnson2016mimic} and PROMOTE \cite{promote}. 
The experimental results reveal that \modelname\ significantly improves KG representation, 
with an average relative performance gain of 12.15\% on MIMIC-III and 9.66\% on PROMOTE across multiple evaluation metrics.
\modelname\ also outperforms the traditional baselines that directly model the high-dimensional sparse EHR attributes without using KGs. These results highlight the effectiveness of \modelname\ in integrating patient-specific context into KG representations, enabling more accurate and tailored predictions in healthcare applications.  
Additionally, case studies show that by integrating external contexts, \modelname\ can learn to adjust the representations of entities and relations in KG, potentially improving the quality and real-world utility of the KG. 

%% file: sections/3_relatedwork.tex
\section{Related Work}

\subsection{Biomedical entity linking.}
Linking entities is a crucial part of building comprehensive healthcare knowledge graphs and is essential for entity predictions. Researchers have proposed conventional methods focusing on setting string-matching rules \cite{d2015ELRule} and leveraging constructed thesauri \cite{aronson2010ELDICT}, and machine learning-based methods \cite{wang2023ELLM}\cite{zhang2022ELML}\cite{ji2020ELLM}\cite{xu2020ELLM} transforming biomedical concepts from raw text into embeddings which are then used to compute similarity scores via distance functions (\textit{e.g.} cosine similarity), and rerank the candidate entities. Recently, some works have explored large language models (LLMs) for biomedical entity linking \cite{xie2024promptlink} due to their unprecedentedly rich prior knowledge as well as language capabilities \cite{lu2023hiprompt}\cite{bhasuran2025preliminary}\cite{singhal2022LLM}\cite{xie2025kerap}. 
Building on these advances, we incorporate entity linking techniques into a unified hypergraph framework for downstream predictive tasks.

\subsection{KG Representations.}
There are many methods aiming to embed entities and relations into vector spaces in a specific dimension. TransE tries to generate embeddings that minimize the distance between the head and the tail entities in a relation, which is one of the easiest methods to train \cite{bordes2011TransE}. The ComplEx embedding uses the Hermitian dot product to produce the embeddings and seperates the embeddings into imaginary and real number parts to handle a large variety of binary relations \cite{trouillon2016ComplEx}. Composition-based Multi-Relational Graph Convolutional Network (CompGCN) extends the traditional Graph Convolutionary Network to include the relations between entities during the message-passing process. The embeddings are generated by aggregating the neighboring entities as well as the relations between entities \cite{vashishth2020CompGCN}. 
These models effectively learn static embeddings over KGs, but none incorporate external patient contexts during embedding learning; our method advances KG representation by embedding both KG structure and patient-specific information within the same hypergraph.

\subsection{Personalized KG for Healthcare.}
Our contextualized KGs are distinct from personalized KGs in several ways.
Unlike personalized KGs that focus on creating various small KGs to support the information needs of different users \cite{jianggraphcare}\cite{xu2023PersonalizedKG}\cite{ye2021PersonalizedKG}\cite{Theo2023PersonalizedKG}\cite{carvalho2023PersonalizedKG}\cite{gao2023drknow}, through knowledge contextualization, we add external contexts into KG while still modeling it as a whole. This approach can achieve the following unique effects distinct from personalized KGs: 
(1) it enables global modeling of interactions between knowledge and contexts, allowing learning to happen across users; 
(2) it utilizes the contexts to adjust knowledge representations in the whole KG, improving the quality and utility of knowledge that can potentially benefit other related KG applications;
(3) it leverages knowledge in the whole KG to improve the modeling of contexts and subsequently improve the modeling of specific users.
Due to the often different application scenarios of personalized KGs compared to our approach, as well as the often complicated and ad hoc processes of creating personalized KGs, we do not empirically compare with personalized KGs in this work and leave this direction as future work.
Whereas personalized KGs isolate subgraphs per user, our unified hypergraph contextualization preserves KG coherence and enables collective learning across patients, demonstrating clear methodological and practical advantages.

%% file: sections/4_method.tex
\section{Method}

\subsection{Preliminaries}
\subsubsection{Notations.} All mathematical notations used in this paper are summarized in a comprehensive table in Section \ref{app:notation_table} of the Supplemental Materials. 

\subsubsection{Knowledge Graph.} A knowledge graph is a multi-relational graph $\mathcal{KG}=(X,R,RT)$, where $X$ denotes the set of entities (nodes), $R$ denotes the set of relations (edges), and $RT\in X\times R \times X$ denotes the relational triples in the KG. 
In other words, knowledge in $\mathcal{KG}$ is stored as a collection of relational triples $RT$. 
For example, the relation triple \textit{(Melatonin, Cause, Dry skin)} indicates the fact that melatonin can cause dry skin.

\subsubsection{Knowledge Graph Representation.} The objective of KG representation learning is to create the mapping functions $\mathcal{F}_X: X \mapsto \mathcal{Z}_X \in \mathbb{R}^{d_{KG}}$ and $\mathcal{F}_R: R\mapsto \mathcal{Z}_R \in \mathbb{R}^{d_{KG}}$ that transform the entities and relations in $\mathcal{KG}$ into a low-dimensional vector space. These representations, $\mathcal{Z}_X$ and $\mathcal{Z}_R$, are designed to capture and preserve the structural and attribute details of the original graph $\mathcal{KG}$, ensuring that the knowledge within the graph is retained.

\subsubsection{Contextualizing KG Representation with EHR for Precision Healthcare.} We study the KG contextualization problem in the context of patient EHR data. Let $p$ represent a patient, and the associated EHR data can be represented as a sequence of time-stamped medical records: $\rho_{p}=\{r_1, r_2, \dots, r_\tau\}$, where $r_i$ represents the $i$-th record associated with the patient, and $\tau$ is the total number of events for the patient. 
Each medical record $r_i$ can be described as a tuple: $r_i=(t_i, a_i, lt_i)$, where $t_i\in \mathbb{R}^{+}$ is the timestamp of the record (\textit{i.e.} when the record is charted), $a_i$ is the medical attribute (\textit{e.g.} diagnosis such as heart failure, prescription such as desmopressin), and $lt_i$ is the optional literal value of $a_i$ (\textit{e.g.} N/A for heart failure, 0.05mg for desmopressin). 
The medical attribute $a_i$ of $r_i$ can be linked to one entity $X_k$ in $\mathcal{KG}$ using entity linking methods. Using the knowledge graph representation learning method $\mathcal{F}_X$, we can obtain the KG representations $(Z_{X_1}, \cdots, Z_{X_\tau})$ of a patient's related KG entities $(X_1, \cdots, X_\tau)$. Consequently, a patient's overall representation $Z_p$ is defined as $Z_p = \mathcal{F_{Z}}(Z_{X_1}, \cdots, Z_{X_\tau})$, where $\mathcal{F_Z}$ is the contextualization function that contextualizes the general factual KG representation into patient-specific knowledge representation.

\subsubsection{Hypergraph.} A hypergraph $\mathcal{HG} = \{V, E\}$ consists of a finite set of vertices (nodes) $V$ and hyperedges $E$, where each hyperedge $e \in E$ is a non-empty subset of the vertex set $V$. Unlike regular graphs where edges connect two vertices, hyperedges can connect any number of vertices (\textit{e.g.} $e_1=\{v_2, v_4, v_7\}$). The size of a hyperedge $e$ is the number of vertices it connects, while the degree $deg_{v}$ of a vertex $v$ is the number of hyperedges it participates in. In our \modelname\ framework, we use hypergraphs to model patient contexts and knowledge, with hyperedges representing patient visits and vertices representing medical attributes.

\subsection{\modelname\ Framework}
\label{sec:framework}
\begin{figure*}[htbp]
\centering
\includegraphics[width=1\linewidth]{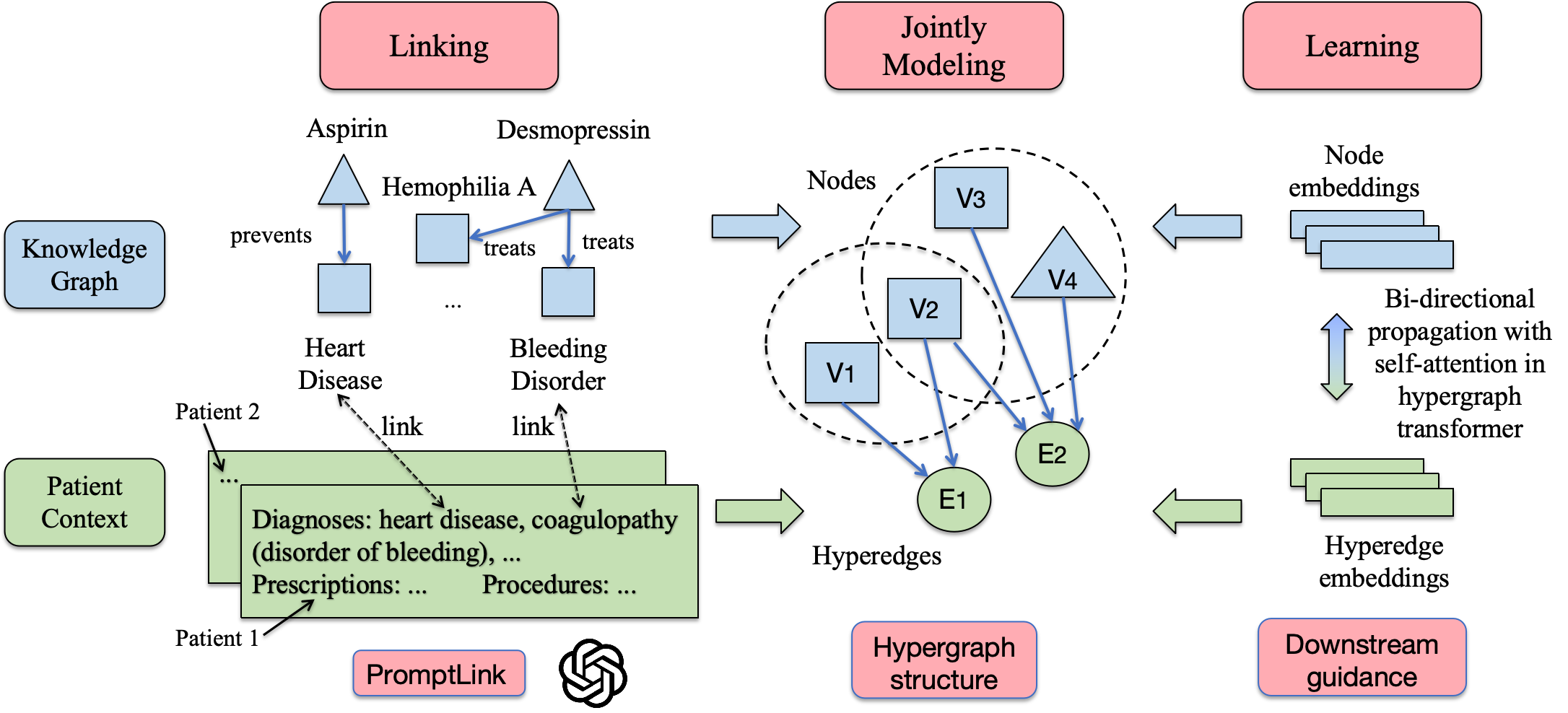}
\caption{Overview of \modelname\ framework. \textit{Left:} Linking knowledge and contexts. \textit{Middle:} Jointly modeling knowledge (e.g., node $V_1$) and contexts (e.g., hyperedge $E_1$, which includes nodes $V_1$ and $V_2$; hyperedge $E_2$, which includes nodes $V_2$, $V_3$, and $V_4$). \textit{Right:} Learning to integrate knowledge and contexts.}
\label{fig:method}
\vspace{-10pt}
\end{figure*}

\modelname\ is a framework designed to contextualize KG representation with patient-specific context for precision healthcare, and the pipeline is described in Figure \ref{fig:method}. First, \modelname\ connects KG entities with relevant context information from EHR by linking medical entities between them (described in Section \ref{sec:Context_Identification}). Then, \modelname\ jointly represents KG knowledge and contextual information from EHR in a hypergraph structure, capturing key relationships between patients, features, and other patients (described in Section \ref{sec:Context_Representation}). Finally, the node and hyperedge embeddings in the hypergraph structure are learned and optimized for downstream precision healthcare tasks (described in Section \ref{sec: Context Injection}). 
This framework requires the availability of contextual data such as EHRs, reliable mappings between patients and their clinical features, and a robust embedding or entity linking model to initialize node representations effectively. These components ensure that context is accurately integrated into the KG and can be leveraged for improved predictive performance.

\subsection{Linking Knowledge and External Contexts}
\label{sec:Context_Identification}
To connect KG-extracted knowledge with relevant patient context, we leverage the advanced LLM-based entity-linking technique - PromptLink \cite{xie2024promptlink}, to accurately link patient-specific context from EHR with KG entities. 
The process begins with preprocessing the contextual medical attribute $a$ in EHR $\rho_{p}$ (\textit{e.g.} diagnosis names, medication names, procedure names) by lowercasing and removing punctuation, ensuring uniformity across both EHR and KG entities. 
We then leverage a pre-trained language model (SAPBERT \cite{liu2020sapbert}) to generate embeddings \( Z_{lm, a} \in \mathbb{R}^{d_{lm}} \) for EHR entity \( a \) and \( Z_{lm, X_k} \in \mathbb{R}^{d_{lm}} \) for KG entity \( X_k \). 
Next, we compute the cosine similarity \( SIM \in [0, 1] \) between the embeddings \( Z_{lm, a} \) and \( Z_{lm, X_k} \), identifying the top-\( LC \) candidates (where \( LC = 10 \)) with the highest similarity scores as potential matches for further analysis. 
At the final stage, we employ a LLM (GPT-4), to establish the entity linking by analyzing the semantic meanings of the candidates. Using both the embeddings and the biomedical knowledge stored in the LLM, we generate final predictions for linking $LK$ (between KG entities $X_k$ and EHR medical attribute $a$). 
This multi-step approach, combining deep embeddings and LLM-based reasoning, ensures accurate entity linking, thereby enabling the integration of patient-specific data into the KG for contextualized healthcare applications. The process can be formalized as:
\begin{equation}
\label{equ:promptlink}
    LK(a, X_k) = \text{PromptLink}(a, X_k).
\end{equation}

\subsection{Jointly Modeling Knowledge and Contexts}
\label{sec:Context_Representation}
After connecting KG knowledge and contextual information from EHR, \modelname\ jointly represents them in a hypergraph structure.
In this hypergraph model, medical attributes such as diagnosis and medications are treated as nodes \(V\). A patient's individual visit $\rho_p$ (\textit{i.e.} single encounter) is represented as a hyperedge \(e\), which connects the relevant nodes corresponding to the medical attributes from that visit. The collection of these hyperedges forms the hyperedge set \(E\) within the hypergraph. Unlike standard graphs that link only two nodes, hypergraphs allow hyperedges to connect multiple nodes at once, making it easier to model the relationship within the patient context. This hypergraph-based approach effectively represents the rich contextual information in EHR data and captures high-order interactions between patients and medical features, providing a more structured and comprehensive representation.

\subsection{Learning to Integrate Knowledge and Contexts}
\label{sec: Context Injection}
\modelname\ learns to integrate knowledge and contexts. To facilitate this process, we initialize node embeddings within the hypergraph to incorporate KG information and patient context. For each node \( v \), associated with an EHR attribute \( a \) and connected to a KG entity \( X_k \), the initial node embedding \( \mathcal{Z}_{v}^{0} \in \mathbb{R}^{d_{KG}} \) is derived from the corresponding KG representation.

We design the learning process inspired by hypergraph transformers \cite{lee2019settransformer}\cite{chien2021settransformer}\cite{xu2023hypergraph}. 
The structures consist of $L$ layers for message passing. In each layer, the embeddings are updated as follows: hyperedge embeddings \( \mathcal{Z}_{e}^{l} \) of layer $l$ are obtained by aggregating node information from layer $l-1$ within each hyperedge \( e \) using aggregation function $f_{V \to E}$, and then the node embeddings \( \mathcal{Z}_{v}^{l} \) are updated by aggregating information from hyperedges connected to the node \( v \) using aggregation function $f_{E \to V}$. These message-passing processes are iteratively applied across layers:
\begin{equation}
\label{equ:hg_propagation}
    \mathcal{Z}_{e}^{l} = f_{V \to E}(\mathcal{Z}_{v}^{l-1}), \quad \mathcal{Z}_{v}^{l} = f_{E \to V}(\mathcal{Z}_{e}^{l})
\end{equation}
Additionally, in the first layer, the hyperedge embedding is aggregated using the initial node embedding $ \mathcal{Z}_{v}^{0}$.

To implement the message-passing functions $f_{V \to E}$ and $f_{E \to V}$, we apply multi-head attention. Here we use notation $S$ to represent a set, which can be a hyperedge or a node. Specifically, if $S$ represents a hyperedge, the set $S$ consists of its connected nodes, and then the embedding $Z_S$ is updated using the embeddings of its connected nodes from the previous layer. Conversely, if $S$ represents a node, the set $S$ consists of its connected hyperedges, and its embedding $Z_S$ is updated using the embeddings of its connected hyperedges from the previous layer. Given an embedding matrix \( \mathcal{Z}_{S} \in \mathbb{R}^{|S| \times d_{hi}} \), the calculation of output 
 $\mathcal{Z'}_{S}$ is defined as:
\begin{equation}
\label{equ:attention0}
    \mathcal{Z'}_{S} = \parallel_{i=1}^{H} \text{Attention}_{i}(Z_S),
\end{equation}
\begin{equation}
\label{equ:attention1}
    \text{Attention}_{i}(Z_S) = \text{softmax} \left( \frac{W_{Q, i}^{l} (Z_S.W_{K, i}^{l})^\top}{\sqrt{d_{hi}/H}} \right) Z_S.W_{V, i}^{l},
\end{equation}
where $W_{Q, i}^{l} \in \mathbb{R}^{1 \times \lfloor d_{hi}/H \rfloor}$, $W_{K, i}^{l} \in \mathbb{R}^{d_{hi} \times \lfloor d_{hi}/H \rfloor}$, $W_{V, i}^{l}  \in \mathbb{R}^{d_{hi} \times \lfloor d_{hi}/H \rfloor}$ are learnable weight matrices in layer $l$, \( H \) is the number of attention heads, and \( d_{hi} \) is the hidden layer dimension. 

It is crucial to capture three key relationships within the patient context: (1) Patient-Attribute Relations – linking patients to their respective attributes; (2) Attribute-Attribute Relations – illustrating how attributes influence each other within a patient; and (3) Patient-Patient Relations – identifying similarities between patients with comparable medical histories or conditions. Our \modelname, which utilizes a hypergraph structure with patient-level hyperedges and attribute-level nodes, facilitates the capture of Patient-Attribute Relations through node-to-hyperedge and hyperedge-to-node propagation, as described in Eq. \ref{equ:hg_propagation}. Additionally, the multi-head attention mechanism, defined in Eqs. \ref{equ:attention0} and \ref{equ:attention1}, enables the model to measure Attribute-Attribute Relations via attention between nodes within a hyperedge, and Patient-Patient Relations through attention between hyperedges connected to a common node.

To guide learning, we define downstream tasks (\( \mathcal{TA} \)) and divide the hypergraph data into training, validation, and test sets for robust evaluation. 
During training, KG information is contextualized using patient data, capturing intricate relationships like patient-attribute, attribute-attribute, and patient-patient relations. 
The resulting final representation, \( \mathcal{Z}_{final} \), incorporates both patient-specific context and KG-derived knowledge, and can then be utilized in various healthcare applications. Since we leverage supervised classification tasks for guidance, the final representation \( \mathcal{Z}_{final} \) is passed to a two-layer fully-connected layer (with size $d_{final}$=48) and the loss function is the binary cross-entropy.

%% file: sections/5_experiment.tex
\section{Experiments}
\subsection{Experimental Settings}
\label{sec:setting}
\subsubsection{KG Dataset.}
We use a large-scale public knowledge graph, iBKH \cite{su2023ibkh}, as the primary KG dataset. iBKH integrates data from various biomedical knowledge bases, offering a comprehensive resource with 2,384,501 entities across 11 categories, including drugs, diseases, symptoms, genes, and pathways. Moreover, with over 48 million relation triples, iBKH facilitates deeper insights into complex biological interactions. More details about how we utilize the KG dataset to generate embeddings are described in Section \ref{app:implementation_details} of Supplemental Materials.

\subsubsection{EHR Datasets.}
We contextualize the knowledge graph by integrating patient-specific data from two EHR datasets: MIMIC-III \cite{johnson2016mimic} and PROMOTE \cite{promote}. Overview of the two EHR datasets are described in Table \ref{tab:data_stat}. MIMIC-III, a public dataset, includes over 53,000 patient records from critical care units at Beth Israel Deaconess Medical Center between 2001 and 2012. 
From MIMIC-III, we follow the settings of Xu et al. \cite{xu2023hypergraph}, to keep 12,353 patient visits (\textit{i.e.} encounters) with 7,423 key medical attributes, including 846 diagnoses, 4,525 prescriptions, and 2,032 procedures. 
For the MIMIC-III dataset, we perform phenotyping prediction, framed as a multi-label classification task. This involves predicting the presence of 25 acute health conditions (e.g., chronic kidney disease) during a patient's future visits based on their current ICU stay records.
PROMOTE is a private dataset containing records of 7,780 stroke patients treated from 2012 to 2021. For this dataset, we extract 2,595 medical attributes, including 1,480 ICD-10 diagnosis codes and 1,115 prescribed medications recorded up to the patients' discharge after their index stroke. 
For the PROMOTE dataset, we use the occurrence of post-stroke cognitive impairment (PSCI) as the label and define it as a binary classification task. 
More details about the two EHR datasets are described in Section \ref{app:ehr_dataset_details} of Supplemental Materials.

\begin{table}[htbp!]
    \centering
    \caption{Overview of the EHR datasets. Dx denotes diagnosis, Rx denotes prescriptions, and Px denotes procedures.}
    \label{tab:data_stat}
    \resizebox{0.45\linewidth}{!}{
    \begin{tabular}{cccc}
    \toprule
    \textbf{EHR} & \textbf{Patient} & \textbf{List of} & \textbf{Task} \\
     \textbf{Dataset} & \textbf{Encounter Count} & \textbf{Features} & \textbf{Count} \\
    \midrule
    MIMIC-III & 12,353 & Dx, Rx, Px & 25 \\
    PROMOTE   & 7,780  & Dx, Rx     & 1 \\
    \bottomrule
    \end{tabular}
    }
\end{table}

\subsubsection{Evaluation Metrics.} In our experiments, we evaluate model performance using key metrics for classification tasks: Accuracy, Area Under the Receiver Operating Characteristic Curve (AUROC), Area Under the Precision-Recall Curve (AUCPR), and the Macro-F1 score. For the binary classification performance on PROMOTE, we follow the default way to compute these metrics. 
For the multilabel classification performance on MIMIC-III, all metrics are macro-averaged to ensure that performance is evaluated equally across all labels.
The model's final performance is determined based on the highest AUROC score achieved on the validation set, as AUROC is a comprehensive metric for evaluating the classification model's effectiveness, and the corresponding test set performance from that epoch is then reported. To ensure robustness, we repeat the experiments five times and report the average performance across these runs. These metrics provide a thorough assessment of the model's predictive capabilities.

\subsubsection{Baseline Methods.}
In our experiments, we compare the downstream risk prediction performance of the following embedding approaches, all of which are patient representations for downstream prediction tasks:
\begin{itemize}[leftmargin=10pt, itemsep=0pt, topsep=0pt, parsep=0pt]
    \item Binary Embedding: A traditional patient risk factor encoding approach~\cite{EHRBinary1}\cite{EHRBinary2}\cite{EHRBinary3} that represents patients using one-hot binary embeddings for each medical attribute. This results in a high-dimensional binary embedding with over one thousand dimensions (7423 for MIMIC-III and 2595 for PROMOTE), where each dimension indicates the presence or absence of a specific medical attribute during a patient's visit.
    \item KG Embedding: This method leverages KG information by mean-pooling KG representations based on the linked patient's medical attributes during a visit. While we use PromptLink~\cite{xie2024promptlink} to perform the entity linking between EHR attributes and KG entities, our approach extends beyond PromptLink by generating and aggregating the KG embeddings into patient-level representations. The resulting 128-dimensional vector captures the patient's relevant KG context in a compact form, offering a lower-dimensional alternative to traditional high-dimensional binary feature embeddings.
    \item \modelname\ Embedding: This 128-D embedding is generated using our proposed method \modelname, which integrates patient-specific contextual information with the KG information. 
\end{itemize}
Each of these embeddings is evaluated using several machine learning methods: Logistic Regression (LR), Support Vector Machine (SVM), Multi-layer Perceptron (MLP), Random Forest (RF), and XGBoost (XGB). Additionally, for the \modelname\ embedding, we also assess the performance during the training process using the patient context integration structure - hypergraph transformer (Hypergraph), since the contextualization process itself also uses the same downstream prediction task. 
These models allow us to thoroughly assess the effectiveness of the different embeddings in capturing the relevant features for healthcare tasks.
The experiments for downstream prediction (Section \ref{sec:main_experiment}) are designed to demonstrate that contextualizing KG embeddings with patient data yields substantially better risk‐prediction performance.  Moreover, our ablation studies (Section \ref{sec:ablation}) show that \modelname\ remains robust under variations in hypergraph construction, entity‐linking strategies, and other design choices.  

\subsubsection{Implementation Details.} Implementation details of the experiments are described in Section \ref{app:implementation_details} of Supplemental Materials. These details include the KG embedding generation method, hypergraph modeling hyperparameters, and baseline ML models' hyperparameter configurations.

\subsection{Performance on Downstream Prediction}
\label{sec:main_experiment}
\begin{table*}[htbp]
    \centering
    \caption{Performance comparison of different embeddings on the downstream risk prediction task. For each evaluation metric, the best performance is highlighted in bold, and the second-best result is underlined.}
    \label{tab:main_results}
    \resizebox{\textwidth}{!}{%
    \begin{tabular}{ccccccccccc}
    \hline
\multirow{2}{*}{\textbf{Embedding}} & \multirow{2}{*}{\textbf{Method}} & \multicolumn{4}{c}{\textbf{MIMIC-III}} & \multicolumn{4}{c}{\textbf{PROMOTE}} \\ \cmidrule(lr){3-6} \cmidrule(lr){7-10}
  &  & \textbf{Accuracy(\%)} & \textbf{AUROC(\%)} & \textbf{AUCPR(\%)} & \textbf{Macro-F1(\%)} & \textbf{Accuracy(\%)} & \textbf{AUROC(\%)} & \textbf{AUCPR(\%)} & \textbf{Macro-F1(\%)} \\  \hline
Binary& LR & 78.58±0.18 & 81.23±0.14 & 68.22±0.32 & 45.79±0.44 & 74.08±0.69 & 60.56±1.96 & 30.67±1.91 & 53.84±1.37 \\ 
Embedding& SVM & 80.41±0.10 & 83.21±0.23 & 71.29±0.37& 41.77±0.04 & 77.42±0.69 & 63.13±1.20 & 33.48±1.61 & 44.28±0.26 \\ 
 & MLP & 76.05±0.11 & 78.09±0.22 & 63.74±0.25 & 44.21±0.22 & 69.30±0.58 & 57.04±1.13 & 28.22±1.92 & 53.46±0.71 \\ 
 & RF & 77.17±0.25 & 81.22±0.11 & 67.45±0.40 & 23.60±0.21 & 77.40±0.11 & 64.73±1.33 & 32.77±1.55 & 43.80±1.12 \\ 
 & XGB & 79.62±0.20 & 82.72±0.15 & 70.83±0.21 & 49.06±0.34 & 77.11±0.91 & 65.01±1.70 & 33.30±2.42 & 49.84±1.17 \\ \hline
KG& LR & 77.72±0.22 & 80.66±0.09 & 66.20±0.36 & 35.80±0.34 & 77.40±0.82 & 61.62±1.70 & 29.79±2.41 & 44.08±0.34 \\ 
Embedding & SVM & 78.38±0.22 & 80.84±0.19 & 67.49±0.39 & 35.40±0.30 & 77.49±0.83 & 54.97±1.39 & 24.49±0.94 & 43.78±0.26 \\ 
 & MLP & 72.08±0.21 & 73.12±0.21 & 53.79±0.54 & 42.47±0.39 & 68.77±1.11 & 57.64±1.55 & 24.05±6.48 & 53.39±1.81 \\ 
 & RF & 76.57±0.17 & 78.98±0.16 & 64.36±0.56 & 25.13±0.95 & 77.59±0.83 & 56.49±1.69 & 25.79±1.53 & 43.75±0.26 \\ 
 & XGB & 75.35±0.13 & 77.27±0.19 & 61.77±0.51 & 38.65±0.49 & 77.48±0.87 & 59.04±1.31 & 28.95±1.63 & 46.21±0.84 \\ \hline
\modelname & LR & 80.32±0.29 & 83.98±0.45 & \underline{72.84±0.91} & 46.28±0.54 & \underline{77.62±0.74} & \underline{66.97±0.51} & \underline{35.65±1.56} & 47.58±0.95 \\ 
Embedding & SVM & 80.51±0.16 & 83.07±0.35 & 71.59±0.62 & 42.83±0.24 & \textbf{77.70±0.90} & 58.07±1.22 & 28.49±1.93 & 44.06±0.54 \\ 
 & MLP & 74.61±0.18 & 77.10±0.30 & 62.15±4.47 & \underline{46.91±0.62} & 69.17±1.32 & 58.72±1.59 & 32.19±1.61 & \textbf{54.60±1.52} \\ 
 & RF & \underline{80.54±0.23} & \underline{84.01±0.22} & 72.80±0.47 & 41.72±0.29 & 74.26±7.92 & 63.74±1.15 & 32.19±1.92 & 46.98±4.74 \\ 
 & XGB & 78.86±0.21 & 81.89±0.61 & 69.25±0.63 & \textbf{47.12±0.63} & 77.02±0.52 & 64.22±1.28 & 32.65±2.85 & 49.43±0.82 \\ 
 & Hypergraph & \textbf{80.86±0.1}7 & \textbf{84.26±0.17} & \textbf{73.12±0.3}7 & 45.30±0.49 & 77.06±0.62 & \textbf{67.37±1.02} & \textbf{35.74±1.67} & \underline{53.28±1.54} \\ \hline
    \end{tabular}%
    }
\end{table*}

Table \ref{tab:main_results} summarizes the downstream risk prediction results on the two datasets to show the impact of contextualization. In this table, our proposed contextualized \modelname\ Embedding consistently outperforms other baseline methods. Across various evaluation metrics on both datasets, the \modelname\ Embedding consistently achieves the best and second-best performance. This performance gain demonstrates the effectiveness of integrating KG information with patient context to generate richer, more representative embeddings. Notably, the hypergraph model within the \modelname\ framework achieves the strongest results across most metrics. This finding suggests that the \modelname\ model alone is powerful enough to provide robust representations and the hypergraph model has a strong modeling ability, eliminating the need to extract embeddings and apply them to external machine learning models. 

The advantages of contextualization within \modelname\ are further emphasized when compared with the KG Embedding. Across all evaluation metrics and on both datasets, the \modelname\ Embedding consistently achieves better results, demonstrating the effectiveness of incorporating contextual information and KG information.
Moreover, we observe an average relative improvement of 12.15\% on MIMIC-III and 9.66\% on PROMOTE by comparing all methods utilizing \modelname\ Embedding and those using KG Embedding.
This improvement highlights the significant impact of contextualization, allowing for a deeper understanding of patient information and stronger performance in downstream tasks.

In comparison to \modelname, the other two methods, Binary Embedding and KG Embedding, show comparable performance in certain models, such as Binary Embedding with SVM. However, both embeddings suffer from significant limitations. Binary Embedding, a traditional approach for generating patient representations in healthcare tasks, faces the curse of dimensionality, especially in medical datasets where the number of attributes can exceed 1,000. This leads to an exponential increase in computational complexity. For instance, when using SVM with Binary Embedding on the MIMIC-III dataset, the computation time exceeded 10 hours—a time cost that is over 20 times longer than the time required by \modelname\ for the same task with SVM. This inefficiency highlights the unsustainable nature of Binary Embedding for large-scale applications. On the other hand, KG Embedding reduces the representation size and also incorporates patient information with KG information. However, the method used for combining them (simple mean-pooling)lacks the sophistication required to capture the intricate relationships between the knowledge graph and patient contexts. As a result, the performance of KG Embedding lags behind that of both Binary Embedding and \modelname\ Embedding. The main limitation here is that the KG information is not sufficiently contextualized, leading to a less effective representation. This is where \modelname\ excels, as it effectively integrates KG and contextual information, modeling intricate relations between them and producing better global representations efficiently for downstream applications.

\subsection{Ablation Studies}
\label{sec:ablation}
\modelname\ consists of three key components: linking, joint modeling, and learning, leveraging EHR data to contextualize KG representations. To validate its design, we conduct comprehensive ablation studies. These studies evaluate the impact of entity linking (Supplemental Materials Section \ref{app:impact_linking_methods}), KG embedding generation (Supplemental Materials Section \ref{app:kg_embedding_generation}), joint modeling of KG and EHR (Supplemental Materials Section \ref{app:jointly_modeling}), hypergraph design (Supplemental Materials Section \ref{app:hypergraph_evaluation}), and learning hyperparameters (Supplemental Materials Section \ref{app:hyperparameter}). 
These ablation studies confirm the effectiveness of \modelname's implementation.

\subsubsection{Entity Linking Methods.}
We evaluate the impact of different entity linking methods on \modelname, comparing PromptLink with BM25, BioBERT, and shuffled embeddings. Results show that PromptLink consistently achieves the best performance. Meanwhile, \modelname\ remains robust with only minor performance drops when using simpler or noisier linking methods. This highlights both the effectiveness of PromptLink and the resilience of \modelname\ to linking errors in real-world settings. Details of the experiments are described in Section \ref{app:impact_linking_methods} of Supplemental Materials.

\subsubsection{KG Embedding Generation Methods.}
We evaluate the impact of different KG embedding methods on \modelname's performance. Details of the experiments are described in Section \ref{app:kg_embedding_generation} of Supplemental Materials. Results show that our default setting (ComplEx + Large) achieves the best overall performance, while alternative methods like TransE and CompGCN perform slightly worse. Overall, \modelname\ remains robust across different embedding configurations, highlighting its adaptability and efficiency in practical applications.

\subsubsection{Jointly Modeling EHR and KG.}
To assess the importance of jointly modeling EHR and KG information, we compare \modelname\ against two ablation baselines: KG Only and EHR Only, as demonstrated in Section \ref{app:jointly_modeling} of Supplemental Materials. Results across both datasets show that \modelname\ significantly outperforms the baselines, demonstrating the complementary value of KG embeddings and EHR-derived hypergraph structures. This confirms the effectiveness of our joint modeling design in enhancing contextualization and predictive accuracy.

\subsubsection{Hypergraph Design.} We evaluate \modelname's hypergraph design by comparing it with state-of-the-art hypergraph models and alternative message-passing strategies. Details and results are described in Section \ref{app:hypergraph_evaluation} of Supplemental Materials. In experiments, \modelname\ consistently outperforms baselines across key metrics, demonstrating its effectiveness in integrating structural and semantic information from both EHRs and KGs. These results validate our contextualization approach and highlight \modelname’s superior scalability and domain-specific design for precision healthcare.

\subsubsection{Hyperparameter Studies.}
We conduct a comprehensive hyperparameter study to assess the impact of key settings on \modelname's performance, as described in Section \ref{app:hyperparameter} of Supplemental Materials. Results on both MIMIC-III and PROMOTE datasets show that our selected configurations achieve strong and stable outcomes, with learning rate, hypergraph transformer depth, hidden size, and attention heads all influencing performance. Among these, the learning rate plays the most critical role in balancing model optimization.

\subsection{Case Studies on Knowledge Representation}

To demonstrate that the representations generated by \modelname\ are effectively contextualized, we compare the knowledge embeddings before and after training. For each KG medical entity pair (Entity A, Entity B), we analyze them from three key perspectives:
(1) Similarity: We compute the cosine similarity between the knowledge representations of each entity pair both before and after training. By comparing these values, we observe an increase in similarity after the application of \modelname\ for most pairs (more details analyzed in Section \ref{app:sim} of Supplemental Materials), indicating that the contextual relationships between entities are better captured. From this analysis, we identify twenty top entity pairs (including pairs of diagnosis and prescription) that show a great increase in representation similarity within both the MIMIC-III and PROMOTE datasets for a case study. Since the cosine similarity values are within the range $[-1, 1]$, the increases in similarity for these listed pairs are significant as they approach 1.
(2) Occurrence and Prevalence: We further investigate the frequency with which each entity pair co-occurs in the same patient context. Additionally, we determine the prevalence of each entity in these contexts by dividing the number of co-occurrences by the total number of occurrences for each entity individually (i.e., Prevalence A and Prevalence B). This allows us to understand how common it is for both entities to appear together and how strongly their occurrence is correlated within a patient's data.
(3) Relation Analysis: To provide further insights into the relationships between the entities, we consult with clinicians for qualitative assessments and cite relevant resources.

\begin{table*}[htbp!]
\centering
\caption{Case studies on knowledge representation. ``Increase'' denotes the increase in similarity between entity pairs after applying \modelname. ``Occurrence'' indicates the number of times the entity pair co-occurs in the same patient context. ``Prevalence A'' and ``Prevalence B'' represent the ratio of co-occurrences to the total occurrences for each entity individually. ``References'' denote existing literature that discusses the relationship between the entity pairs.}
\label{tab:case}
\resizebox{1\linewidth}{!}{
\begin{tabular}{ccccccccc}
\hline
\textbf{Dataset and Pair} & \textbf{ID} & \textbf{Entity A} & \textbf{Entity B} & \textbf{Increase} & \textbf{Occurrence} & \textbf{Prevalence A} & \textbf{Prevalence B} & \textbf{References} \\ \hline
MIMIC-III & I & Benign neonatal seizures & Pneumonia & 1.08 & 25 & 67.57\% & 40.98\% & \cite{case1} \\ 
Diagnosis & II & Benign neonatal seizures & Congenital hemolytic anemia & 1.01 & 30 & 81.08\% & 34.88\% & \cite{case2} \\ 
& III & Upper extremity deep vein thrombosis &   Limb injury & 1.01 & 4 & 57.14\% & 100.00\% &  \cite{case3}\\
& IV & Congenital hemolytic anemia &  Pneumonia & 1.00 & 38 & 44.19\% & 62.30\% &  \cite{case4}\\ 
& V & Closed fracture of bones, unspecified & Pneumothorax & 0.99 & 43 & 36.44\% & 78.18\% & \cite{case5} \\ \hline

MIMIC-III & VI & Starter PN D5 & Heparin sodium & 1.21 & 7 & 53.85\% & 58.33\% & \cite{case6} \\
Prescription & VII & Insulin & Acetaminophen & 1.13 & 5566 & 75.34\% & 69.38\% & \cite{case7} \\
& VIII & Drug withdrawal syndrome neonatal & Indomethacin sodium & 1.12 & 9 & 36.00\% & 81.82\% & \cite{case8} \\
& IX & Syringe (Neonatal) & Indomethacin sodium & 1.11 & 11 & 44.00\% & 84.62\% & \cite{case9}\\ 
& X & Urethral stricture & Pulmonary tuberculosis & 1.11 & 19 & 79.17\% & 63.33\% & \cite{case10} \\ \hline

PROMOTE & XI & Respiratory system disease & Medical device complication & 1.09 & 1043 & 66.01\% & 68.21\% &\cite{case11} \\ 
Diagnosis & XII & Human immunodeficiency virus infectious disease & HIV infection & 1.07 & 95 & 75.40\% & 84.07\% & \cite{case12}\\ 
& XIII & Symptoms of nervous \& musculoskeletal systems & Long term (current) drug therapy & 1.06 & 2990 & 72.89\% & 55.12\% & \cite{case13}\\ 
& XIV & Essential hypertension & Chronic fatigue syndrome & 1.02 & 2966 & 44.37\% & 89.61\% & \cite{case14}\\ 
& XV & Essential hypertension & Cerebrovascular disease & 0.99 & 4282 & 64.06\% & 87.57\% & \cite{case15}\\ \hline

PROMOTE & XVI &  Isopropyl alcohol & Lancets & 0.92 & 95 & 98.96\% & 50.00\% & \cite{case161720}\\ 
Prescription & XVII & Blood glucose & Lancets & 0.87 & 181 & 88.73\% & 95.26\% & \cite{case161720}\\
& XVIII  & Emtricitabine \& Tenofovir disoproxil fumarate& Ritonavir & 0.87 & 12 & 75.00\% & 66.67\% & \cite{case18}\\ 
& XIX & Mycophenolate mofetil & Tacrolimus & 0.83 & 59 & 71.08\% & 74.68\% & \cite{case19}\\ 
& XX & Blood glucose & Isopropyl alcohol & 0.76 & 86 & 54.78\% & 89.58\% & \cite{case161720} \\ \hline
\end{tabular}
}
\end{table*}

The case study results are described in Table \ref{tab:case}. 
Overall, \modelname\ enhances the similarity between entity pairs that are similar in patient contexts, even when their relationships are not explicitly captured by the KG, showing the power of our contextualization. For instance, in Case I,  ``Benign neonatal seizures'' and ``Pneumonia'' are two distinct diagnostic conditions that, according to the KG, have no direct causal link. However, in the EHR context, these dignoses are associated with the medical attributes ``Convulsions in newborn'' and ``Pneumonia'', which often co-occur in newborns. Pneumonia, for example, may indirectly contribute to neonatal seizures by causing physiological stress that makes seizures more likely in affected infants \cite{case1}. While the KG might not capture this indirect relationship, the contextual co-occurrence in patient data suggests a meaningful association.
Another example is seen in Case VII, which involves the prescription entity pair of Insulin and Acetaminophen. These two drugs do not have a direct pharmacological interaction, as insulin is primarily used to manage blood sugar levels in diabetic patients, while acetaminophen is typically prescribed for pain or fever relief. However, they are linked to their corresponding medical attributes within the EHR context and frequently co-occur. This can be explained by the phenomenon that patients with diabetes often require both medications—insulin for managing their condition and acetaminophen to alleviate common symptoms like pain or fever \cite{case7}.  Although the KG does not reflect a direct connection between these drugs, their frequent co-prescription in clinical practice highlights a contextual relationship, which \modelname\ is able to capture.
In summary, this case study analysis demonstrates that \modelname\ contextualizes the KG knowledge and generates better representations.

%% file: sections/6_conclusion.tex
\section{Conclusion}
In this study, we propose \modelname, a novel framework that addresses the critical need for contextualized knowledge representations for precision healthcare by integrating context-specific information from EHRs with general knowledge from KGs. 
In this framework, we first use entity-linking techniques to connect general KG knowledge with patient contexts from EHRs, and then apply a hypergraph model to contextualize the knowledge via hyperedges. Finally, we employ hypergraph transformers guided by downstream prediction tasks to jointly learn contextualized representations for both KG and patients, leveraging the combined knowledge.
In experiments, \modelname\ shows significant improvements in healthcare prediction tasks across multiple evaluation metrics. Additionally, by integrating external contexts, \modelname\ can learn to adjust the representations of
entities and relations in KG, potentially improving the quality and real-world utility of knowledge. 
The application of \modelname\ allows for the effective contextualization of knowledge and advances precision healthcare, capturing the intricate relationships between patient data and general knowledge in a way that traditional KGs fail to do. 
 

%% file: sections/7_acknowledgement.tex

\section*{Supplemental Material Statement}
The source code and dataset download instructions are available on the GitHub page: \href{https://github.com/constantjxyz/HypKG}{https://github.com/constantjxyz/HypKG}. Additional resources—including the notation table, algorithm descriptions, dataset details, implementation notes, and ablation studies—are provided in the Supplementary Materials of the extended manuscript available on arXiv: \href{https://arxiv.org/abs/2507.19726}{https://arxiv.org/abs/2507.19726}.

%% file: supplementary.tex
\newpage
\appendix
\section*{Supplemental Materials}
\section{Notation Table}
\label{app:notation_table}

\begin{table}[htbp!]
\centering
\caption{Notations used in the paper.}
\label{tab:notations}
\resizebox{0.7\linewidth}{!}{
\begin{tabular}{cl}
\hline
\textbf{Notation}       & \textbf{Description}                                                                                     \\ \hline
$\mathcal{KG}$          & Knowledge graph, represented as $(X, R, RT)$.                                                           \\ 
$X$                     & Set of entities (nodes) in the knowledge graph.                                                         \\ 
$R$                     & Set of relations (edges) in the knowledge graph.                                                        \\ 
$RT$                    & Relational triples in the knowledge graph.                          \\ 
$\mathcal{F}_X$         & Mapping function for KG entities.               \\ 
$\mathcal{F}_R$         & Mapping function for KG relations.              \\ 
$\mathcal{Z}_X$         & Vector representation of entities in $X$.                                               \\ 
$\mathcal{Z}_R$         & Vector representation of relations in $R$.                                              \\ 
$d_{KG}$ & Dimension of KG representation ($\mathcal{Z}_X$ and $\mathcal{Z}_R$).\\
$p$                     & A patient.                                                                                              \\ 
$\rho_p$                & Sequence of time-stamped medical records for a patient $p$.                                             \\ 
$r_i$                   & The $i$-th medical record of a patient $p$.                        \\ 
$t_i$                   & Timestamp of the $i$-th medical record.                                                                 \\ 
$a_i$                   & Medical attribute in the $i$-th medical record (e.g., diagnosis).                \\ 
$lt_i$                   & Optional literal value associated with $a_i$.                                                           \\ 
$X_k$                   & Knowledge graph entity linked to EHR medical attribute $a$.                                               \\ 
$\mathcal{Z}_{X_{k}}$         & Representation of a KG entity $X_k$.                                               \\ 
$Z_p$                   & KG representation of a patient $p$             \\ 
$\mathcal{F_Z}$         &  Contextualization function for patient embedding $Z_p$. \\
$\tau$                  & Total number of medical records for a patient.                                                          \\ 
$\mathcal{HG}$          & Hypergraph, represented as $(V, E)$.                                                                    \\ 
$V$                     & Set of vertices (nodes) in the hypergraph.                                                              \\ 
$E$                     & Set of hyperedges in the hypergraph.                                                                    \\ 
$v$ & A (vertex) node in the hypergraph.\\
$e$                     & A hyperedge, connecting multiple nodes in $V$.                                                          \\ 
$deg_v$ & Degree of a vertex $v$. \\

$\mathcal{Z}_{lm, a}$         & Language-model-encoded embedding of an EHR medical attribute $a$.                                            \\ 
$\mathcal{Z}_{lm, X_k}$         & Language-model-encoded embedding of an EHR medical attribute $a$.                                            \\ 
$d_{lm}$ & Dimension of a language-model-encoded embedding.\\
$SIM$ & Cosine similarity score between embeddings. \\
$LK$           & Linking function between EHR attributes and KG entities. \\
$LC$ & Number of top candidates for entity linking, $LC=10$. \\
$\mathcal{Z}_v^l$         & Embedding of a node $v$ in the hypergraph's $l$-th layer. \\ 
$\mathcal{Z}_e^l$         & Embedding of a hyperedge $e$ in the hypergraph's $l$-th layer.                                             \\ 
$\mathcal{Z}_v^0$         & Initial embedding of a node $v$ in the hypergraph. \\ 
$f_{V \to E}$           & Aggregation function to update hyperedge embeddings.                               \\ 
$f_{E \to V}$           & Aggregation function to update node embeddings.                               \\ 
$S$ & Input set (nodes in a hyperedge or hyperedges connected to a node) \\
$\mathcal{Z}_S$	& Embedding matrix of a set $S$. \\
$\mathcal{Z}'_S$	& Updated embedding matrix of a set $S$ after message passing. \\
$W_{Q, i}^{l}$ & Learnable weight matrix in layer $l$.\\
$W_{K, i}^{l}$ & Learnable weight matrix in layer $l$.\\
$W_{V,i}^{l}$ & Learnable weight matrix in layer $l$. \\
$L$ & Total number of message-passing layers in the hypergraph transformer.\\
$d_{hi}$ & Size of the hidden state in hypergraph transformer.\\
$H$                     & Number of attention heads in the hypergraph transformer.                                                \\ 
$\mathcal{TA}$          & Downstream healthcare prediction tasks.                                                                 \\ 
$\mathcal{Z}_{final}$   & Final representation for downstream tasks.                          \\ 
$d_{final}$             & Dimension of the final representation for downstream tasks. \\
$lr$ & Learning rate of the learning process.\\
$\mathcal{KG'}$ & Subsampled knowledge graph derived from the original $\mathcal{KG}$. \\
$X'$ & Entity set in the subsampled knowledge graph $\mathcal{KG'}$.\\
$RT'$ & Relation triple set in the subsampled knowledge graph $\mathcal{KG'}$.\\
$K$ & Retained relations when generating KG representation, $K=800$.\\
\hline
\end{tabular}
}
\end{table}

\section{\modelname\ Algorithm}
\label{app:algorithm}
Here we present the pseudocode for \modelname\ to provide readers with a clear understanding of the entire procedure.

\begin{algorithm}[htbp]
   \caption{\textbf{HypKG for Precision Healthcare}}
   \label{alg:hypkg_brief}
   \resizebox{0.9\linewidth}{!}{
   \begin{minipage}{1\linewidth}
   \begin{algorithmic}
   \STATE {\bfseries Input:} Knowledge Graph; Patient EHR data; Hypergraph parameters (e.g., \#layers \(L\), \#heads \(H\));
   \STATE {\bfseries Output:} Contextualized KG embeddings \(\mathcal{Z}_{final}\);
   
   \vspace{0.2cm}
   \STATE {\bfseries Step 1: Linking Knowledge and External Contexts}
   \STATE Preprocess EHR attributes, encode with a pre-trained language model.
   \STATE Retrieve top-\(LC\) candidates from KG by cosine similarity.
   \STATE Use an LLM to select the best match linking \(LK(a, X_k)\) for EHR attribute $a$ and KG entity $X_k$.

   \vspace{0.2cm}
   \STATE {\bfseries Step 2: Jointly Modeling Knowledge and Contexts}
   \STATE Nodes $V$: Each unique medical attribute (linked to a KG entity).
   \STATE Hyperedges $E$: Each patient visit forms one hyperedge that connects multiple nodes, representing relationships among attributes.
   \STATE Initialize node embeddings \(\mathcal{Z}_v^0\) from KG representations.

   \vspace{0.2cm}
   \STATE {\bfseries Step 3: Learning to Integrate Knowledge and Contexts}
   \FOR{$l = 1$ {\bfseries to} $L$}
   \STATE Update hyperedge embedding: \( \mathcal{Z}_e^l \leftarrow f_{V\to E}(\mathcal{Z}_{v}^{l-1}) \).
   \STATE Update node embedding: \( \mathcal{Z}_v^l \leftarrow f_{E\to V}(\mathcal{Z}_{e}^{l}) \).
   \ENDFOR
   \STATE Train on downstream tasks (e.g., classification) using a supervised loss (e.g., cross-entropy).

   \vspace{0.2cm}
   \STATE {\bfseries Return:} Final embeddings \(\{\mathcal{Z}_{final}\}\) and trained model.
   \end{algorithmic}
   \end{minipage}
   }
\end{algorithm}

\section{EHR Dataset Details}
\label{app:ehr_dataset_details}

An overview of the dataset characteristics, including patient encounter counts, available features, and task counts, is provided in Table \ref{tab:data_stat}. More descriptions about the EHR datasets are in the main manuscript's Section 4.1. 

\begin{table}[htbp!]
    \centering
    \caption{Overview of the EHR datasets. Dx denotes diagnosis, Rx denotes prescriptions, and Px denotes procedures.}
    \label{tab:data_stat}
    \resizebox{0.5\linewidth}{!}{
    \begin{tabular}{cccc}
    \toprule
    \textbf{EHR} & \textbf{Patient} & \textbf{List of} & \textbf{Task} \\
     \textbf{Dataset} & \textbf{Encounter Count} & \textbf{Features} & \textbf{Count} \\
    \midrule
    MIMIC-III & 12,353 & Dx, Rx, Px & 25 \\
    PROMOTE   & 7,780  & Dx, Rx     & 1 \\
    \bottomrule
    \end{tabular}
    }
\end{table}

Moreover, to guide the contextualization process, we introduce several classification tasks as supervision signals. For the MIMIC-III dataset, we perform phenotyping prediction, framed as a multi-label classification task. This involves predicting the presence of 25 acute health conditions (e.g., chronic kidney disease) during a patient's future visits based on their current ICU stay records. The full list of phenotypes is provided in Table \ref{tab:task_name}. For the PROMOTE dataset, we use the occurrence of post-stroke cognitive impairment (PSCI) as the label and define it as a binary classification task.
\begin{table}[htbp!]
\centering
\caption{Full list of downstream prediction tasks utilized in the experiments. ``Prevalence'' represents the percentage of positive samples out of the total samples in the dataset.}
\label{tab:task_name}
\resizebox{0.7\linewidth}{!}{
\begin{tabular}{llc}
\hline
\textbf{Dataset} & \textbf{Phenotype} & \textbf{Prevalence} \\ \hline
MIMIC-III & Acute and unspecified renal failure & 29.3\% \\
& Acute cerebrovascular disease & 7.1\% \\
& Acute myocardial infarction & 7.4\% \\
& Cardiac dysrhythmias & 43.2\% \\
& Chronic kidney disease & 25.5\% \\
& Chronic obstructive pulmonary disease & 18.0\% \\
& Complications of surgical/medical care & 84.0\% \\
& Conduction disorders & 2.4\% \\
& Congestive heart failure; nonhypertensive & 39.2\% \\
& Coronary atherosclerosis and related & 29.6\% \\
& Diabetes mellitus with complications & 36.2\% \\
& Diabetes mellitus without complication & 42.1\% \\
& Disorders of lipid metabolism & 27.6\% \\
& Essential hypertension & 35.4\% \\
& Fluid and electrolyte disorders & 44.4\% \\
& Gastrointestinal hemorrhage & 27.8\% \\
& Hypertension with complications & 59.5\% \\
& Other liver diseases & 21.9\% \\
& Other lower respiratory disease & 35.4\% \\
& Other upper respiratory disease & 9.5\% \\
& Pleurisy; pneumothorax; pulmonary collapse & 33.9\% \\
& Pneumonia & 21.7\% \\
& Respiratory failure; insufficiency; arrest & 32.8\% \\
& Septicemia (except in labor) & 26.5\% \\
& Shock & 12.2\% \\
\hline
PROMOTE & Post-stroke Cognitive Impairment & 22.3\% \\
\hline
\end{tabular}
}
\end{table}

\section{Implementation Details}
\label{app:implementation_details}
\subsection{KG Embedding.}
Considering the extremely large size of the original KG, we first sample entities and relations to facilitate the representation process. After linking EHR context to KG entities (enforce 1-to-1 linking), we subsample a KG $\mathcal{KG'}$ with entity set $X'$ and relation triple set $RT'$. For each entity, we retain the top-$K$ relation triples based on edge degree (the sum of the connected entities' degrees within $\mathcal{KG'}$). The choice of $K$ is 800. Using the \textit{PyKEEN} library \cite{ali2021pykeen}, we employ the ComplEx model to generate KG embeddings with dimension $d_{KG}$=128. More details about the KG embeddings are described in Appendix \ref{app:variants}.

\subsection{Hyperparameter Settings for Hypergraph Model.}
During the learning process, the input dataset is split into training, validation, and test sets with a ratio of $(70\%, 10\%, 20\%)$. We implement our model using PyTorch~\cite{pytorch}, and for optimization, we apply the Adam optimizer~\cite{KingmaB14Adam}. The learning rates $lr$ are set to 1e-3 for the MIMIC-III dataset and 1e-4 for the PROMOTE dataset. Additionally, we use a weight decay of 1e-3 and train the model for 1,000 epochs. The model architecture includes three Set Transformer layers, with an input embedding dimension $d_{KG}$ of 128, a hidden state size $d_{hi}$ of 48, and $H=$4 attention heads. This configuration balances the model’s complexity and its capacity to capture meaningful interactions within the data.

\subsection{Hyperparameter Settings for Other ML Models.}
Hyperparameter settings on LR, SVM, MLP, RF, and XGB for two EHR datasets are described in Table \ref{tab:para_mimic_ml} and \ref{tab:para_promote_ml}. These models are trained with different embeddings using \textit{scikit-learn} \cite{sklearn} and \textit{xgboost} \cite{xgboost} Python libraries.

\begin{table}[htbp]
    \centering
    \caption{Hyperparameter setting details of various machine learning models on MIMIC-III. Unmentioned hyperparameters are kept as the default value of \textit{sklearn} and \textit{xgboost} module.}
    \label{tab:para_mimic_ml}
    \resizebox{0.6\linewidth}{!}{
\begin{tabular}{cl}
\hline
\textbf{Method} & \textbf{Hyperparameters} \\ \hline
LR   & max\_iter=1000, C=1 \\ \hline
SVM  & kernel=`rbf', C=10, gamma=`scale' \\ \hline
MLP  & hidden\_layer\_sizes=(64, 64), activation=`relu', \\
              & batch\_size=`auto', learning\_rate=`constant', \\
              & learning\_rate\_init=1e-3, max\_iter=1000 \\ \hline
RF   & n\_estimators=500, max\_depth=10 \\ \hline
XGB  & max\_depth=10, max\_leaves=10, n\_estimators=50, \\
              & learning\_rate=0.5 \\ \hline
\end{tabular}
    }
\end{table}

\begin{table}[htbp]
    \centering
    \caption{Hyperparameter setting details of various machine learning models on PROMOTE. Unmentioned hyperparameters are kept as the default value of \textit{sklearn} and \textit{xgboost} modules.}
    \label{tab:para_promote_ml}
    \resizebox{0.6\linewidth}{!}{
\begin{tabular}{cl}
\hline
\textbf{Method} & \textbf{Hyperparameters} \\ \hline
LR   & max\_iter=1000, C=1 \\ \hline
SVM  & kernel=`rbf', C=10, gamma=`scale' \\ \hline
MLP  & hidden\_layer\_sizes=(64, 64), activation=`relu', \\
              & batch\_size=`auto', learning\_rate=`constant', \\
              & learning\_rate\_init=1e-3, max\_iter=1000 \\ \hline
RF   & n\_estimators=1000, max\_depth=20 \\ \hline
XGB  & max\_depth=10, max\_leaves=10, n\_estimators=100, \\
              & learning\_rate=0.1 \\ \hline
\end{tabular}
    }
\end{table}

\section{Ablation Study of Entity Linking Methods}
\label{app:impact_linking_methods}

Entity-linking is a critical step in the \modelname\ framework, as it connects patient-specific EHR context to KG entities. PromptLink, as demonstrated in its original study \cite{xie2024promptlink}, is considered a state-of-the-art (SOTA) method for concept linking, and is leveraged in our \modelname\ framework. 

To optimize the original PromptLink pipeline, we have made several technical adjustments to balance accuracy and computational efficiency. Concepts with the same name or matched to the same UMLS concept are directly linked using the UMLS tool to reduce cost. For unmatched concepts, we adapt the pipeline and prompts to enforce one-to-one linking via an LLM, ensuring that every concept is linked to its closest match to facilitate the contextualization step.

To evaluate the impact of entity linking methods and \modelname's sensitivity to entity linking methods and errors (e.g., wrong linking pairs), we replace PromptLink with simpler linking methods that introduce more linking errors and compare the performance. Specifically, we conduct experiments using the following approaches:
\begin{itemize}
    \item BM25: A classical term-based retrieval model based on token matching and ranking \cite{robertson2009BM25}.
    \item BioBERT: A pre-trained transformer model tailored for biomedical text representation and retrieval \cite{lee2020biobert}.
    \item Shuffling: Shuffle the generated KG embeddings randomly that introduce more linking errors.
\end{itemize}

We evaluate each linking method using the MIMIC-III and PROMOTE datasets. All other experimental settings, tasks, and configurations remain consistent. Table~\ref{tab:impact_linking_methods} summarizes the results, highlighting the performance of \modelname\ with each method.

\begin{table*}[htbp]
\centering
\caption{Ablation study results of changing different entity linking methods.}
\label{tab:impact_linking_methods}
\resizebox{\textwidth}{!}{%
\begin{tabular}{ccccccccc}
\hline
\multirow{2}{*}{\textbf{Method}} & \multicolumn{4}{c}{\textbf{MIMIC-III}} & \multicolumn{4}{c}{\textbf{PROMOTE}} \\ \cmidrule(lr){2-5} \cmidrule(lr){6-9}
  & \textbf{Accuracy(\%)} & \textbf{AUROC(\%)} & \textbf{AUCPR(\%)} & \textbf{Macro-F1(\%)} & \textbf{Accuracy(\%)} & \textbf{AUROC(\%)} & \textbf{AUCPR(\%)} & \textbf{Macro-F1(\%)} \\  \hline
\modelname\ with BM25          & 80.25 $\pm$ 0.26       & 83.48 $\pm$ 0.48    & 71.78 $\pm$ 0.61    & 41.89 $\pm$ 2.69       & 76.50 $\pm$ 0.39       & 66.13 $\pm$ 1.57    & 34.77 $\pm$ 1.22    & 50.27 $\pm$ 1.19       \\ 
\modelname\ with BioBERT       & 80.26 $\pm$ 0.16       & 83.63 $\pm$ 0.14    & 71.85 $\pm$ 0.34    & 42.55 $\pm$ 0.57       & 76.74 $\pm$ 0.49       & 66.08 $\pm$ 0.64    & 33.37 $\pm$ 0.78    & 47.30 $\pm$ 0.70       \\ 
\modelname\ with shuffling     & 80.18 $\pm$ 0.28       & 83.15 $\pm$ 0.27    & 71.29 $\pm$ 0.36    & 42.01 $\pm$ 0.71       & 76.46 $\pm$ 0.45       & 65.42 $\pm$ 0.58    & 34.05 $\pm$ 0.78    & 51.50 $\pm$ 2.69       \\ 
Ours   & \textbf{80.86 $\pm$ 0.17} & \textbf{84.26 $\pm$ 0.17} & \textbf{73.12 $\pm$ 0.37} & \textbf{45.30 $\pm$ 0.49} & \textbf{77.06 $\pm$ 0.62} & \textbf{67.37 $\pm$ 1.02} & \textbf{35.74 $\pm$ 1.67} & \textbf{53.28 $\pm$ 1.54} \\ 
\bottomrule
\end{tabular}%
}
\end{table*}

The results show that \modelname\ achieves the best performance with PromptLink as the entity-linking method, consistently outperforming other compared methods across all metrics and datasets. This further proves why PromptLink is a suitable linking method for \modelname and validates our choice. Moreover, the relatively small performance gap (1\%–3\%) demonstrates that \modelname\ is not overly sensitive to the entity linking method, remaining robust even with simpler linking methods and many wrong linking pairs. This robustness ensures that \modelname\ can be reliably applied in real-world scenarios where linking methods vary in accuracy and computational complexity, maintaining strong performance despite these variations.

\section{Ablation Study of KG Embedding Generation Methods}
\label{app:kg_embedding_generation}

KG embeddings are utilized to initialize the hypergraph modeling and learning process for \modelname. We evaluate the impact of different KG embeddings on \modelname\ by experimenting with alternative KG embedding generation techniques. Specifically, we replace ComplEx with other widely used methods, such as TransE and CompGCN, to assess their effects on performance. Additionally, we adjust the relation filter threshold parameter $K$ (as discussed in the KG Embedding section in Appendix \ref{app:implementation_details}) from 800 (the ``Large'' setting) to 80 (the ``Small'' setting). All other experimental settings, tasks, and configurations remain consistent.

The results, summarized in Table \ref{tab:KGE}, show that our default configuration (ComplEx + Large) consistently achieves the best performance across both datasets, validating our choice of KG embedding for the experiments. As for the other compared methods, TransE and CompGCN perform slightly worse than ComplEx. Furthermore, CompGCN presents significant computational challenges. The running time for the ``CompGCN + Large'' setting exceeds 10 days per epoch on our KG dataset, which is why we omit this setting from the experiment. Interestingly, variations in KG embedding methods and relation filter thresholds do not significantly affect the overall performance of \modelname, indicating its robustness. While certain settings, such as changing the KG embeddings or reducing the number of relations, may introduce slight disadvantages, most of the useful relations in the KG are still effectively extracted. \modelname's architecture remains flexible and resilient enough to maintain strong performance. This robustness is crucial for practical applications, as it allows \modelname\ to adapt to different KG embedding techniques without sacrificing significant accuracy or efficiency.

\begin{table*}[htbp]
    \centering
    \caption{Ablation study results of changing different KG embedding generation methods.}
    \label{tab:KGE}
    \resizebox{\textwidth}{!}{%
    \begin{tabular}{ccccccccc}
    \hline
\multirow{2}{*}{\textbf{Embedding}} & \multicolumn{4}{c}{\textbf{MIMIC-III}} & \multicolumn{4}{c}{\textbf{PROMOTE}} \\ \cmidrule(lr){2-5} \cmidrule(lr){6-9}
& \textbf{Accuracy(\%)} & \textbf{AUROC(\%)} & \textbf{AUCPR(\%)} & \textbf{Macro-F1(\%)} & \textbf{Accuracy(\%)} & \textbf{AUROC(\%)} & \textbf{AUCPR(\%)} & \textbf{Macro-F1(\%)} \\  \hline
ComplEx + Large & \textbf{80.87$\pm$0.17} & \textbf{84.26$\pm$0.17} & \textbf{73.11$\pm$0.37} & 45.43$\pm$0.49 & 77.06$\pm$0.62 & \textbf{67.37$\pm$1.02} & \textbf{35.74$\pm$1.67} & \textbf{53.28$\pm$1.54} \\ 
ComplEx + Small & 80.85$\pm$0.12 & 84.17$\pm$0.15 & 73.01$\pm$0.46 & 45.40$\pm$0.30 & \textbf{77.48$\pm$0.23} & 66.72$\pm$1.26 & 35.06$\pm$2.58 & 49.86$\pm$2.21 \\
TransE + Large & 80.86$\pm$0.24 & 84.16$\pm$0.24 & 72.88$\pm$0.48 & 44.88$\pm$0.68 & 76.98$\pm$0.32 & 67.21$\pm$1.46 & 35.50$\pm$2.32 & 52.28$\pm$2.32 \\
TransE + Small & 80.84$\pm$0.15 & 84.14$\pm$0.17 & 72.94$\pm$0.39 & 45.16$\pm$0.52 & 77.30$\pm$0.33 & 67.08$\pm$0.75 & 35.41$\pm$2.34 & 48.45$\pm$2.10 \\ 
CompGCN + Small & 80.80$\pm$0.29 & 84.19$\pm$0.23 & 72.92$\pm$0.32 & \textbf{45.50$\pm$0.36} & 77.12$\pm$0.53 & 66.82$\pm$0.96 & 35.20$\pm$2.41 & 51.94$\pm$1.83 \\ \hline
    \end{tabular}%
    }
\end{table*}

\section{Ablation Study of Jointly Modeling EHR and KG}
\label{app:jointly_modeling}

\modelname\ jointly models EHR information (patient-specific context) and KG information in the hypergraph structure. To evaluate the significance of this jointly modeling design, we conduct ablation experiments comparing \modelname\ with the following baselines:
\begin{itemize}
    \item KG Only: KG embeddings are utilized to initialize the hypergraph model, but the hyperedge structures (i.e., which nodes are contained in a hyperedge) are randomly initialized (by randomly shuffling the original hyperedge structure). 
    \item EHR Only: KG embeddings are randomly initialized without any prior information from KGs, but the hyperedge structure is obtained from the EHR dataset. 
\end{itemize}
These baselines are designed to isolate the contributions of KG embeddings and hypergraph structures derived from EHR data. All other experimental settings, tasks, and configurations remain consistent. By comparing these configurations with \modelname, we can quantify the impact of jointly modeling these two components. 

Table~\ref{tab:impact_kg} presents the results of this ablation study, evaluated on two datasets, MIMIC-III and PROMOTE, across multiple performance metrics: Accuracy, AUROC, AUCPR, and Macro-F1. 
In the table, \modelname\ significantly outperforms both KG Only and EHR Only baselines across all metrics. The results highlight the complementary roles of EHR and KG data in improving predictive performance. The KG Only baseline suffers from random hyperedge structures, while the EHR Only baseline lacks the external knowledge provided by KG embeddings. In contrast, \modelname's integration of these two sources achieves superior contextualization and robustness. 

These findings emphasize the importance of our joint modeling design in \modelname, showcasing its ability to effectively utilize both EHR and KG information for improved contextualization and predictive accuracy.

\begin{table*}[htbp]
\centering
\caption{Ablation study results of jointly modeling.}
\label{tab:impact_kg}
\resizebox{1\textwidth}{!}{%
\begin{tabular}{ccccccccc}
\hline
\multirow{2}{*}{\textbf{Method}} & \multicolumn{4}{c}{\textbf{MIMIC-III}} & \multicolumn{4}{c}{\textbf{PROMOTE}} \\ \cmidrule(lr){2-5} \cmidrule(lr){6-9}
  & \textbf{Accuracy(\%)} & \textbf{AUROC(\%)} & \textbf{AUCPR(\%)} & \textbf{Macro-F1(\%)} & \textbf{Accuracy(\%)} & \textbf{AUROC(\%)} & \textbf{AUCPR(\%)} & \textbf{Macro-F1(\%)} \\  \hline
KG Only & 71.39 $\pm$ 0.21 & 69.29 $\pm$ 0.16 & 53.11 $\pm$ 0.24 & 16.49 $\pm$ 0.69 & 75.42 $\pm$ 4.10 & 53.27 $\pm$ 2.96 & 24.50 $\pm$ 2.41 &46.24 $\pm$ 1.79 \\
EHR Only        & 76.85 $\pm$ 1.09       & 78.43 $\pm$ 1.70    & 65.01 $\pm$ 1.14    & 42.55 $\pm$ 0.48       & 75.37 $\pm$ 0.17       & 56.33 $\pm$ 1.25    & 27.40 $\pm$ 1.34    & 43.62 $\pm$ 0.05       \\ 
\modelname\ (Ours)                 & \textbf{80.86 $\pm$ 0.17} & \textbf{84.26 $\pm$ 0.17} & \textbf{73.12 $\pm$ 0.37} & \textbf{45.30 $\pm$ 0.49} & \textbf{77.06 $\pm$ 0.62} & \textbf{67.37 $\pm$ 1.02} & \textbf{35.74 $\pm$ 1.67} & \textbf{53.28 $\pm$ 1.54} \\ 
\bottomrule
\end{tabular}%
}
\end{table*}

\section{Ablation Study of Hypergraph Design}
\label{app:hypergraph_evaluation}

Hypergraph-based models are a powerful approach for capturing high-order relationships in complex datasets, such as EHRs. In this paper, we leverage the hypergraph structure to contextualize input representations by incorporating both structural and semantic information. This section evaluates the performance of \modelname\ against other SOTA hypergraph models, validating its contextualization design.

We benchmarked \modelname\ against prominent hypergraph models, including HGTN \cite{li2023hgtn}, HyperGCN \cite{yadati2019hypergcn}, HCHA \cite{bai2021hcha}, and HypEHR \cite{xu2023hypergraph}, using the MIMIC-III dataset. This publicly available benchmark dataset was selected to facilitate reproducibility and enable fair comparisons across different approaches. These models leverage hypergraph structures to capture intricate relationships among patients, medical codes, and other clinical entities.

As presented in Table~\ref{tab:hypergraph_comparison}, \modelname\ consistently demonstrates superior performance across key metrics, including Accuracy, AUROC, and AUCPR, while maintaining competitive Macro-F1 scores. Notably, HGTN achieves the highest Macro-F1 score, but its single-task design requires training 25 separate models to handle each classification task individually. In contrast, \modelname\ integrates all tasks into a unified framework, offering significantly improved scalability and efficiency without compromising accuracy. The results highlight the robustness of \modelname, which balances predictive power with the ability to handle imbalanced data. By leveraging patient-specific information and external knowledge from KGs, \modelname\ achieves SOTA performance in contextualized healthcare predictions, validating our technical design of contextualization.

\begin{table*}[htbp]
\centering
\caption{Performance comparison of hypergraph models on MIMIC-III.}
\label{tab:hypergraph_comparison}
\resizebox{0.6\textwidth}{!}{%
\begin{tabular}{ccccc}
\toprule
\textbf{Method}         & \textbf{Accuracy(\%)} & \textbf{AUROC(\%)} & \textbf{AUCPR(\%)} & \textbf{Macro-F1(\%)} \\ \midrule
HGTN                   & 77.66 $\pm$ 0.21       & 72.16 $\pm$ 0.11    & 51.05 $\pm$ 0.33    & \textbf{58.63 $\pm$ 0.21}       \\ 
HyperGCN               & 78.01 $\pm$ 0.23       & 80.34 $\pm$ 0.15    & 67.68 $\pm$ 0.16    & 39.29 $\pm$ 0.20       \\ 
HCHA                   & 78.07 $\pm$ 0.28       & 80.42 $\pm$ 0.17    & 68.56 $\pm$ 0.15    & 37.78 $\pm$ 0.22       \\ 
HypEHR                 & 79.07 $\pm$ 0.31       & 82.19 $\pm$ 0.13    & 71.08 $\pm$ 0.17    & 41.51 $\pm$ 0.25       \\ 
\modelname (Ours)             & \textbf{80.86 $\pm$ 0.17} & \textbf{84.26 $\pm$ 0.17} & \textbf{73.12 $\pm$ 0.37} & 45.30 $\pm$ 0.49       \\ 
\bottomrule
\end{tabular}%
}
\end{table*}

To refine the contextualization capabilities of hypergraph models, we further explored alternative design choices, including incorporating graph neural network (GNN) projections and adding KG-related nodes directly into hyperedges. These methods aim to enhance the structural and semantic representation of clinical data. As shown in Table~\ref{tab:hypergraph_transformer}, \modelname\ consistently outperforms these variants. Our design, which initializes the hypergraph transformer with high-quality KG embeddings, proves most effective. This approach ensures that \modelname\ captures both global biomedical knowledge and patient-specific context, leading to improved predictions for EHR tasks.

\begin{table*}[htbp]
\centering
\caption{Performance comparison of hypergraph message passing methods.}
\label{tab:hypergraph_transformer}
\resizebox{0.7\textwidth}{!}{%
\begin{tabular}{ccccc}
\toprule
\textbf{Method}         & \textbf{Accuracy(\%)} & \textbf{AUROC(\%)} & \textbf{AUCPR(\%)} & \textbf{Macro-F1(\%)} \\ \midrule
\modelname (Ours)             & \textbf{80.86 $\pm$ 0.17} & \textbf{84.26 $\pm$ 0.17} & \textbf{73.12 $\pm$ 0.37} & \textbf{45.30 $\pm$ 0.49} \\ 
Adding GNN             & 80.12 $\pm$ 0.08       & 83.38 $\pm$ 0.17    & 71.69 $\pm$ 0.31    & 43.14 $\pm$ 0.71       \\ 
Adding KG-related Nodes & 80.32 $\pm$ 0.29      & 83.60 $\pm$ 0.36    & 71.94 $\pm$ 0.62    & 42.54 $\pm$ 0.48       \\ 
\bottomrule
\end{tabular}%
}
\end{table*}

Besides validating our hypergraph design, it is essential to highlight the differences between \modelname\ and some key referenced models. HypEHR \cite{xu2023hypergraph} focuses solely on EHR data, using hypergraphs to model high-order relationships among medical codes and patient visits. It generates node embeddings based solely on the graph structures, without incorporating the semantic meaning of clinical concepts. AllSet \cite{chien2021settransformer} and SetTransformer \cite{lee2019settransformer}, on the other hand, are general-purpose hypergraph frameworks designed to model multiset interactions efficiently. These models are typically used for node classification on hypergraphs and lack the domain-specific tailoring to precision healthcare seen in \modelname.

\section{Hyperparameter Studies}
\label{app:hyperparameter}

We conduct an in-depth study of the effects of different hyperparameters on \modelname's learning process. From the results in Table \ref{tab:para_mimic} and Table \ref{tab:para_promote}, it is evident that our chosen hyperparameter settings are well-validated, consistently showing strong performance across both datasets. Among all the compared hyperparameters, the learning rate ($lr$) plays a critical role in balancing model optimization between overfitting and underfitting. The depth of hypergraph transformer layers ($L$), the size of the hidden state ($d_{hi}$), and the number of attention heads ($H$) also need to be tuned, as they influence the model's complexity.

\begin{table}[htbp]
\centering
\caption{Hyperparameter tuning results on MIMIC-III. Our chosen hyperparameter setting is ``$lr$=1e-3, $L$=3, $d_{hi}$=48, $H$=4''. }
\label{tab:para_mimic}
\resizebox{0.65\linewidth}{!}{
\begin{tabular}{ccccc}
\hline
\textbf{Parameter} & \textbf{Accuracy(\%)} & \textbf{AUROC(\%)} & \textbf{AUCPR(\%)} & \textbf{Macro-F1(\%)} \\ \hline
Ours & \textbf{80.87$\pm$0.17} & \textbf{84.26$\pm$0.17} & \textbf{73.11$\pm$0.37} & 45.43$\pm$0.49 \\ 
$lr$=1e-2 & 80.83$\pm$0.20 & 84.06$\pm$0.11 & 72.72$\pm$0.31 & 45.43$\pm$0.93 \\ 
$lr$=1e-4 & 79.93$\pm$0.20 & 83.05$\pm$0.31 & 71.24$\pm$0.66 & 39.79$\pm$0.58 \\ 
$L$=2 & 80.84$\pm$0.20 & 84.17$\pm$0.22 & 72.98$\pm$0.47 & 45.32$\pm$0.57 \\
$L$=4 & 80.84$\pm$0.22 & 84.19$\pm$0.24 & 72.99$\pm$0.43 & 45.60$\pm$0.50 \\ 
$d_{hi}$=64 & 80.83$\pm$0.17 & 84.22$\pm$0.19 & 72.98$\pm$0.28 & \textbf{45.91$\pm$0.43} \\ 
$d_{hi}$=32 & 80.84$\pm$0.18 & 84.17$\pm$0.21 & 72.96$\pm$0.46 & 44.97$\pm$0.34 \\ 
$head$=2 & 80.83$\pm$0.16 & 84.17$\pm$0.23 & 72.96$\pm$0.33 & 45.19$\pm$0.51 \\
$head$=8 & 80.85$\pm$0.09 & 84.23$\pm$0.20 & 73.08$\pm$0.39 & 45.45$\pm$0.55 \\ \hline

\end{tabular}
}
\end{table}

\begin{table}[htbp]
\centering
\caption{Hyperparameter tuning results on PROMOTE. Our chosen hyperparameter setting is ``$lr$=1e-4, $L$=3, $d_{hi}$=48, $H$=4''. }
\label{tab:para_promote}
\resizebox{0.65\linewidth}{!}{
\begin{tabular}{ccccc}
\hline
\textbf{Parameter} & \textbf{Accuracy(\%)} & \textbf{AUROC(\%)} & \textbf{AUCPR(\%)} & \textbf{Macro-F1(\%)} \\ \hline
Ours & 77.06$\pm$0.62 & \textbf{67.37$\pm$1.02} & \textbf{35.74$\pm$1.67} & \textbf{53.28$\pm$1.54} \\ 
$lr$=1e-3 & 77.22$\pm$0.58 & 67.25$\pm$1.40 & 35.73$\pm$1.84 & 53.26$\pm$1.67 \\ 
$lr$=1e-5 & \textbf{77.76$\pm$0.67} & 63.34$\pm$2.13 & 31.93$\pm$2.35 & 44.80$\pm$0.50 \\ 
$L$=2 & 77.47$\pm$0.92 & 67.08$\pm$1.41 & 35.02$\pm$3.81 & 51.90$\pm$3.26 \\
$L$=4 & 77.43$\pm$0.43 & 66.60$\pm$1.43 & 35.44$\pm$1.77 & 49.93$\pm$1.77 \\ 
$d_{hi}$=64 & 77.21$\pm$0.69 & 67.06$\pm$1.05 & 35.33$\pm$1.66 & 52.61$\pm$1.22 \\ 
$d_{hi}$=32 & 77.26$\pm$0.57 & 66.45$\pm$0.89 & 34.81$\pm$1.79 & 50.97$\pm$2.00 \\
$H$=2 & 77.12$\pm$0.37 & 66.83$\pm$1.21 & 35.20$\pm$2.46 & 52.07$\pm$2.64 \\ 
$H$=8 & 77.01$\pm$0.61 & 66.50$\pm$0.53 & 34.15$\pm$1.83 & 52.39$\pm$3.99 \\ \hline
\end{tabular}
}
\end{table}

\section{\modelname\ Variants}
\label{app:variants}
In the experiments, we use the \textit{PyKEEN} library \cite{ali2021pykeen} to generate KG embeddings with the ComplEx model, which are then concatenated with embeddings from the DeepWalk \cite{agarwal2019walk} model and SAPBERT. This approach incorporates additional contextual information from both topological and semantic features. To reduce dimensionality, we apply Principal Component Analysis (PCA), producing a final 128-D representation. To validate our approach, we also conduct experiments with different variants of \modelname by replacing the concatenated embedding with individual 128-D embeddings, each reduced by PCA. The results, shown in Table \ref{tab:variants}, indicate that the concatenated embedding achieves the best performance, suggesting that the KG information from ComplEx, topological insights from DeepWalk, and semantic information from SAPBERT complement each other effectively.

\begin{table*}[htbp]
    \centering
    \caption{Performance Comparison between Variants of \modelname.}
    \label{tab:variants}
    \resizebox{\linewidth}{!}{
    \begin{tabular}{cccccccccc}
    \hline
\multirow{2}{*}{\textbf{Embedding}} & \multicolumn{4}{c}{\textbf{MIMIC-III}} & \multicolumn{4}{c}{\textbf{PROMOTE}} \\ \cmidrule(lr){2-5} \cmidrule(lr){6-9}
  & \textbf{Accuracy(\%)} & \textbf{AUROC(\%)} & \textbf{AUCPR(\%)} & \textbf{Macro-F1(\%)} & \textbf{Accuracy(\%)} & \textbf{AUROC(\%)} & \textbf{AUCPR(\%)} & \textbf{Macro-F1(\%)} \\  \hline
Ours & \textbf{80.87$\pm$0.17} & \textbf{84.26$\pm$0.17} & \textbf{73.11$\pm$0.37} & \textbf{45.43$\pm$0.49} & 77.06$\pm$0.62 & \textbf{67.37$\pm$1.02} & \textbf{35.74$\pm$1.67} & \textbf{53.28$\pm$1.54} \\ 
KG alone & 80.66$\pm$0.17 & 83.98$\pm$0.19 & 72.63$\pm$0.33 & 44.59$\pm$0.62 & 77.43$\pm$0.82 & 65.58$\pm$1.73 & 33.89$\pm$3.08 & 48.26$\pm$2.48 \\ 
SAPBERT alone & 80.80$\pm$0.17 & 84.18$\pm$0.19 & 72.83$\pm$0.36 & 45.40$\pm$0.41 & \textbf{77.52$\pm$0.38} & 66.99$\pm$1.83 & 35.61$\pm$2.24 & 49.48$\pm$2.26 \\ 
Walk alone & 80.81$\pm$0.27 & 84.14$\pm$0.23 & 73.10$\pm$0.48 & 45.30$\pm$0.50 & 77.06$\pm$0.72 & 66.92$\pm$1.59 & 34.72$\pm$2.09 & 50.08$\pm$2.43 \\ \hline
    \end{tabular}
    }
\end{table*}

\section{Knowledge Representation Similarity Study}
\label{app:sim}
\begin{figure}[htbp]
    \centering
    \includegraphics[width=0.75\linewidth]{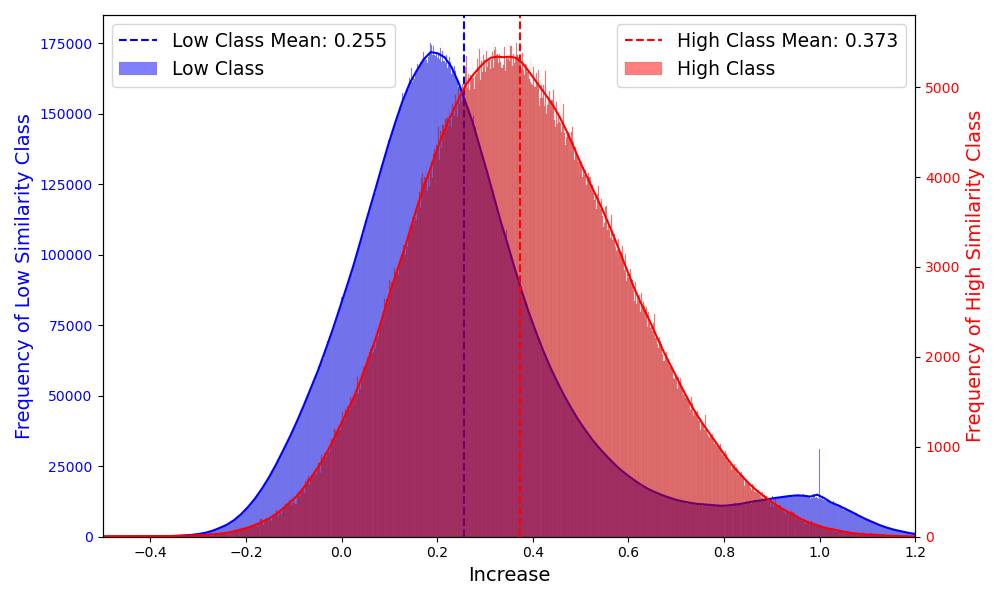}
    \caption{Histogram showing the similarity increase for two different classes after \modelname's learning process in the MIMIC-III dataset.}
    \label{fig:mimic_sim}
\end{figure}

\begin{figure}[htbp]
    \centering
    \includegraphics[width=0.75\linewidth]{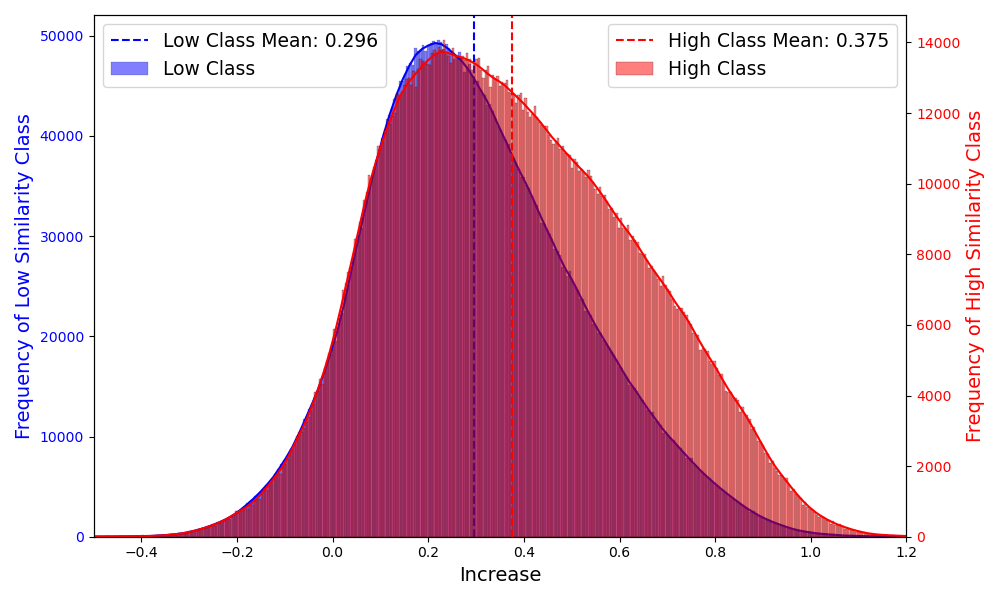}
    \caption{Histogram showing the similarity increase for two different classes after \modelname's learning process in the PROMOTE dataset.}
    \label{fig:promote_sim}
\end{figure}

In this section, we further investigate the increase in the similarity of KG representations after the \modelname\ learning process for different entity pairs. To perform this analysis, we first classify KG entity pairs into two distinct groups based on their linked context information from EHRs:
\begin{itemize}[leftmargin=10pt]
    \item \textit{High Class}: The linked entities appear within the same patient context in the EHR. 
    \item \textit{Low Class}: The linked entities do not appear together in any patient contexts. 
\end{itemize}
Next, we analyze the similarity increase for KG entity pairs within these two classes and present the results in the histograms in Figures \ref{fig:mimic_sim} and \ref{fig:promote_sim}. From these figures, we observe a significant difference between the two classes. Specifically, the similarity increase for \textit{High Class} is substantially greater than that of \textit{Low Class}, with a relative improvement of 46.27\% on MIMIC-III and 26.69\% on PROMOTE, respectively. This difference is attributed to the fact that entity pairs in \textit{High Class} appear within the same patient context, allowing \modelname\ to capture the relationships more effectively within the hypergraph transformer structure.
These results provide strong evidence that \modelname\ successfully contextualizes the knowledge within the KG by leveraging patient-specific context information. 
The ability to enhance similarity among entities that share patient contexts further demonstrates the model's capability to improve the quality of KG representations.

\section{Generalizability of \modelname\ to Other Domains}
\label{sec:app_generalizability}

The underlying framework of \modelname\ is inherently generalizable to other domains that require the contextualization of KGs with domain-specific data. Domains such as e-commerce, user modeling, and recommendation systems often involve scenarios where user-specific contextual data needs to be integrated with external knowledge graphs to enhance predictions and recommendations. In these domains, the core mechanisms of \modelname, including hypergraph-based contextualization and the seamless integration of external context into a unified KG, remain highly applicable.
\begin{itemize}
    \item Hypergraph-Based Contextualization: \modelname\ employs a hypergraph structure to model high-order relationships between entities, attributes, and external context. This approach is versatile and can be adapted to represent interactions such as user-product associations, product-category hierarchies, or user-to-user similarities in domains like e-commerce.
    \item Integration of External Context: The framework integrates KG representation with external contextual information. For example, in user modeling, external contexts such as browsing history, preferences, or engagement patterns can be linked with external KGs like product catalogs or social graphs to provide contextualization information.
\end{itemize}

While the core architecture of \modelname\ is general, its application in other domains may require domain-specific adaptations. For example, applications in other domains need a new domain-specific entity linking method, since the linking method leveraged by \modelname\ is healthcare-oriented, leveraging language models trained on the medical corpus. Besides, the external context also needs to be tailored for specific applications, and the relationships modeled in the hypergraph must align with the domain. 

\modelname\ offers a robust framework for contextualizing knowledge graphs by unifying domain-specific external context and structural knowledge. While healthcare serves as the initial domain of focus, the design principles underlying \modelname\ are flexible and extendable to other domains with similar requirements. Since our focus is on the healthcare application domain, we leave applications on other domains as future work.